\documentclass[journal]{IEEEtran}

\usepackage{amsmath,amssymb,amsfonts}
\usepackage{algorithmic}
\usepackage{algorithm}  
\usepackage{graphicx}
\usepackage{float}
\usepackage{soul}
\usepackage{subfig}
\usepackage{color}
\usepackage{multirow}
\usepackage{fancyhdr}
\usepackage[citecolor=blue, colorlinks]{hyperref} 
\usepackage{cite}

\ifCLASSINFOpdf
\else
\fi
\begin{document}
	
	\title{Semi-Supervised Subspace Clustering via Tensor Low-Rank Representation}
	
    \author{Yuheng~Jia,~Guanxing~Lu,~Hui~Liu,~Junhui~Hou,~\IEEEmembership{Senior~Member,~IEEE}
		\thanks{This work was supported in part by the National Natural Science Foundation of China under Grant 62106044, in part by the Natural Science Foundation of Jiangsu Province under Grant BK20210221, in part by the Hong Kong University Grants Committee under Grant UGC/FDS11/E02/22, in part by the ZhiShan Youth Scholar Program from Southeast University 2242022R40015, and in part sponsored by CCF-DiDi GAIA Collaborative Research Funds for Young Scholars. Corresponding author: Hui Liu.
  
  Y. Jia is with the School of Computer Science and Engineering, Southeast University, Nanjing 210096, China; G. Lu is with the Chien-Shiung Wu College, Southeast University, Nanjing 211102, China; H. Liu is with the School of Computing Information Sciences, Caritas Institute of Higher Education, Hong Kong; J. Hou is with the Department of Computer Science, City University of Hong Kong, Kowloon, Hong Kong (e-mail: yhjia@seu.edu.cn; guanxing@seu.edu.cn; hliu99-c@my.cityu.edu.hk; jh.hou@cityu.edu.hk).} 
	}
	
	\maketitle
 \thispagestyle{fancy}
\lfoot{}
\cfoot{\scriptsize{Copyright © 20xx IEEE. Personal use of this material is permitted. However, permission to use this material for any other purposes must be obtained from the IEEE by sending an email to pubs-permissions@ieee.org.}}
\renewcommand{\headrulewidth}{0mm}
	
	\begin{abstract}
		In this letter, we propose a novel semi-supervised subspace clustering method, which is able to simultaneously augment the initial supervisory information and construct a discriminative affinity matrix.
		By representing the limited amount of supervisory information as a pairwise constraint matrix, we observe that the ideal affinity matrix for clustering shares the same low-rank structure as the ideal pairwise constraint matrix.
		Thus, we stack the two matrices into a 3-D tensor, where a global low-rank constraint is imposed to promote the affinity matrix construction and augment the initial pairwise constraints synchronously.
		Besides, we use the local geometry structure of input samples to complement the global low-rank prior to achieve better affinity matrix learning.
		The proposed model is formulated as a Laplacian graph regularized convex low-rank tensor representation problem, which is further solved with an alternative iterative algorithm.
		In addition, we propose to refine the affinity matrix with the augmented pairwise constraints.
		Comprehensive experimental results on eight commonly-used benchmark datasets demonstrate the superiority of our method over state-of-the-art methods. \textcolor{magenta}{The code is publicly available at https://github.com/GuanxingLu/Subspace-Clustering}.
	\end{abstract}
	
	\IEEEpeerreviewmaketitle

	\section{Introduction}
	
	High-dimensional data are ubiquitously in many areas like image processing, DNA microarray technology, etc.
	The high-dimensional data can often be well approximated by a set of linear subspaces, but the subspace membership of a certain sample is unknown \cite{5714408,peng2022xai}.
	Subspace clustering aims to divide the data samples into different subspaces, which is an important tool to model the high-dimensional data. 
	The state-of-the-art subspace clustering methods \cite{6482137, 6180173} are based on self-expressiveness, which represent high-dimensional data by the linear combination of itself, and enforce a subspace-preserving prior on the self-representation matrix. 
	The representation coefficients capture the global geometric relationships of samples and can act as an affinity matrix, and the subspace segmentation can be obtained by applying spectral clustering on the generated affinity matrix.
	The most well-known subspace clustering methods include sparse subspace clustering \cite{6482137} and low-rank representation \cite{6180173}. 
	
	In many real-world applications, some supervisory information is available, e.g., the label information of a dataset, and the relationships between two samples.
	Generally, those supervisory information can be represented by two kinds of pairwise constraints, i.e., must-link constraints and cannot-link constraints indicating whether two samples belong to the same category or not.
	As the supervisory information is widespread and provides a discriminative description of the samples, many semi-supervised subspace clustering methods \cite{7726045, yang2017semi, 7931596, 9454509, 6750063, 6890188, 7298624, 7968309, 8545800} were proposed to incorporate them.
	Based on the type of the supervisory information, we roughly divide these methods into three classes.
	The first kind of methods include the must-link constraints. 
	For example, \cite{7726045, yang2017semi} incorporated the must-links as hard constraints, which restricted the samples with a must-link to have exactly the same representation.
	The second kind of methods integrate the cannot-link constraints.
	For instance, \cite{7931596} required the affinity relationship of two samples with a cannot-link to be 0.
	Liu \MakeLowercase{\textit{et al.}} \cite{9454509} first enhanced the initial cannot-links by a graph Laplacian term, and then inhibited the affinity of two samples with a cannot-link.
	The third kind of methods can incorporate both the must-links and cannot-links, which generally assume two samples with a must-link should have a higher affinity, while those with a cannot-link should have a lower affinity \cite{6750063, 6890188, 7298624, 7968309, 8545800}.
	However, the above-mentioned semi-supervised subspace clustering methods exploit the supervisory information from a local perspective, but overlook the global structure of the pairwise constraints, which is also important to semi-supervised affinity matrix learning.
	In other words, the previous methods under-use the supervisory information to some extent.
	
	\begin{figure}[!t]
		\centering
		\includegraphics[width=9cm]{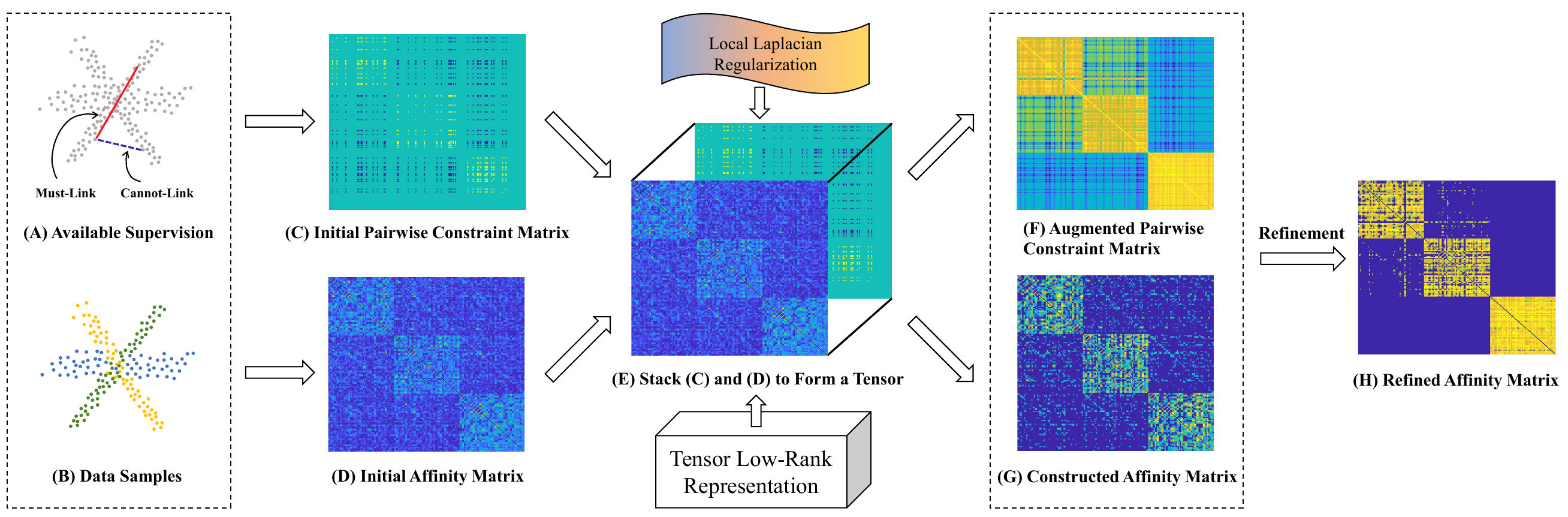}
		\caption{Illustration of the proposed method, which adaptively learns the affinity and enhances the pairwise constraints simultaneously by using their identical global low-rank structure.}
		\vspace{-0.6cm}
		\label{fig_zhiguantu}
	\end{figure}
	To this end, we propose a new semi-supervised subspace clustering method shown in Fig. \ref{fig_zhiguantu}, which explores the supervisory information from a global manner.
	Specifically, in the ideal case, the pairwise constraint matrix is low-rank, as if all the pairwise relationships of samples are available, we could encode the pairwise constraint matrix as a binary low-rank matrix.
	Meanwhile, the ideal affinity matrix is also low-rank, as a sample should only be represented by the samples from the same class.
	More importantly, they share an identical low-rank structure. 
	Based on such an observation, we stack them into a 3-dimensional tensor, and regulate a global low-rank constraint to the formed tensor.
	By seeking the tensor low-rank representation, we can refine the affinity matrix with the available pairwise constraints, and at the same time, augment the initial pairwise constraints with the learned affinity matrix. 
	Besides, we encode the local geometry structure of the data samples to complement the global low-rank prior.
	The proposed model is formulated as a convex optimization model, which can be solved efficiently.
	Finally, we use the augmented pairwise constraint matrix to further refine the affinity matrix. 
	Extensive experiments on 8 datasets w.r.t. 2 metrics demonstrate that our method outperforms the state-of-the-art semi-supervised subspace clustering methods to a large extent.
	
	\section{Preliminary}\label{section_2}
	In this letter, we denote tensors by boldface Euler script letters, e.g., $\mathcal{A}$, matrices by boldface capital letters, e.g., $\mathbf{A}$, vectors by boldface lowercase letters, e.g., $\mathbf{a}$, and scalars by lowercase letters, e.g., $a$. 
	$\|\cdot \|_{2,1}$, $\|\cdot \|_{\infty}$, $\|\cdot\|_F$ and $\|\cdot\|_*$ are the $\ell_{2,1}$ norm, the infinity norm, and the Frobenius norm, and the nuclear norm (i.e., the sum of the singular values) of a matrix.
	$\mathbf{X}\!=\![\mathbf{x}_{1}, \mathbf{x}_{2}, ..., \mathbf{x}_{n}]\in \mathbb{R}^{d\times n}$ is the data matrix, where $\mathbf{x}_{i}\in \mathbb{R}^{d\times1}$ specifies the $d$-dimensional vectorial representation of the $i$-th sample, and $n$ is the number of samples.
	Let $\mathbf{\Omega}_{m}\!=\!\left\{(i, j) \mid \mathbf{x}_{i}\right.$ and $\mathbf{x}_{j}$ belong to the same class$\}$ and $\mathbf{\Omega}_{c}\!=\!\left\{(i, j) \mid \mathbf{x}_{i}\right.$ and $\mathbf{x}_{j}$ belong to different classes$\}$ stand for the available must-link set and cannot-link set.
	We can encode the pairwise constraints as a matrix $\mathbf{B}\in\mathbb{R}^{n\times n}$:
	\begin{equation}
		\mathbf{B}_{ij}=\left\{\begin{aligned}
			1, &~~\text{if}~(i, j) \in \mathbf{\Omega}_{m} \\
			-1, &~~\text{if}~(i, j) \in \mathbf{\Omega}_{c}.
		\end{aligned}\right.
	\end{equation}
	
	Subspace clustering aims to segment a set of samples into a group of subspaces. 
	Particularly, self-expressive-based subspace clustering methods have attracted great attention,  which learn a self-representation matrix to act as the affinity. 
	For example, 
	Liu \MakeLowercase{\textit{et al.}} \cite{6180173} proposed to learn a low-rank affinity matrix by optimizing 
	\begin{equation}
		\min_{\mathbf{Z}, \mathbf{E} }\quad\|\mathbf{Z}\|_{*}+\lambda\|\mathbf{E}\|_{2,1} ~\text { s.t. } \mathbf{X}=\mathbf{X Z}+\mathbf{E},
		\label{eq_LRR}
	\end{equation}
	where $\mathbf{Z}\!\in\!\mathbb{R}^{n\times n}$ is the representation matrix, $\mathbf{E}\in\mathbb{R}^{d\times n}$ denotes the reconstruction error, and $\lambda\!>\!0$ is a hyper-parameter.
	
	Recently, many semi-supervised subspace clustering methods were proposed by incorporating the pairwise constraints \cite{7726045, yang2017semi, 7931596, 9454509, 6750063, 6890188, 7298624, 7968309, 8545800}. How to include the supervisory information is the crux of semi-supervised subspace clustering. \textit{Generally, the existing methods incorporate the pairwise constraints from a local perspective, i.e., expand (resp. reduce) the value of $\mathbf{Z}_{ij}$ if $\mathbf{x}_i$ and  $\mathbf{x}_j$ has a must-link (resp. cannot-link).}

\begin{algorithm}[!t]
		\renewcommand{\algorithmicrequire}{\textbf{Input:}}
		\renewcommand{\algorithmicensure}{\textbf{Output:}}
		\caption{Solve Eq. \eqref{eq_origin} by ADMM}
		\label{alg1}
		\begin{algorithmic}[1]
			\REQUIRE data $\mathbf{X}$, pairwise constraints $\mathbf{\Omega}_{m}$,$\mathbf{\Omega}_{c}$, hyper-parameters $\lambda$,$\beta$. 
			
			\STATE Initialize: $\mathcal{\mathcal{C}}^{(0)}\!=\!\mathcal{Y}_{2}^{(0)}\!=\!0$, $\mathbf{B}^{(0)}\!=\!\mathbf{Z}^{(0)}=\mathbf{D}^{(0)}\!=\!\mathbf{E}^{(0)}\!=\!\mathbf{Y}_{1}^{(0)}\!=\!\mathbf{Y}_{3}^{(0)}\!=\!0, \rho\!=\!1.1, \mu^{(0)}\!=\!1 \mathrm{e}\!-\!3, \mu_{\max }\!=\!1 \mathrm{e} 10$.

			\REPEAT
			\STATE Update $\mathcal{C}$ by $\mathcal{C}^{(k+1)}=\mathcal{S}_{\frac{1}{\mu^{(k)}}}(\mathcal{M}^{(k)}+\mathcal{Y}_{2}^{(k)}/\mu^{(k)})$, where $\mathcal{S}$ is the tensor singular value thresholding operator \cite{8606166};
			\STATE Update $\mathbf{Z}$ by $\mathbf{Z}^{(k+1)}\!=\!\left(\mathbf{I}+\mathbf{X}^{\!\top} \mathbf{X}\right)^{\!-1}(\mathbf{X}^{\!\top}(\mathbf{X}-\mathbf{E}^{(k)})+
		\mathcal{C}^{(k)}(:,:,2)+(\mathbf{X}^{\!\top} \mathbf{Y}^{(k)}_{1}-\mathcal{Y}^{(k)}_{2}(:,:,2))/\mu^{(k)})$;
			\STATE Update $\mathbf{B}$ by $\mathbf{B}^{(k+1)}\!=\!(\mu^{(k)}(\mathcal{C}^{(k)}(:,:,2)\!+\!\mathbf{D}^{(k)})\!-\!(\mathcal{Y}_{2}^{(k)}(:,:,2)\!+\!\mathbf{Y}^{(k)}_{3}))/(\beta(\mathbf{L}\!+\!\mathbf{L}^{\top})\!+\!2\mu^{(k)}\mathbf{I})$;
			\STATE Update $\mathbf{D}$ by 	
			\vspace{-0.3cm}
			\begin{equation}
    		\mathbf{D}^{(k+1)}_{ij}=\left\{\begin{aligned}
    			&s, ~~\text{if} ~(i, j) \in \mathbf{\Omega}_{m} \\
    			&-s, ~~\text{if} ~(i, j) \in \mathbf{\Omega}_{c} \\
    		&\mathbf{B}^{(k)}_{ij}+\mathbf{Y}^{(k)}_{3ij}/\mu^{(k)}, ~~\text {otherwise;}
    		\end{aligned}\right.
    		\nonumber
	        \end{equation}
	        \vspace{-0.3cm}
			\STATE Update $\mathbf{E}$ by 	
			\begin{equation}
        	\mathbf{e}^{(k+1)}_{j}\!=\!\left\{\begin{aligned}
        		&\frac{\left\|\mathbf{q}^{(k)}_{j}\right\|_{2}\!-\!\lambda / \mu^{(k)}}{\left\|\mathbf{q}^{(k)}_{j}\right\|_{2}} \mathbf{q}^{(k)}_{j}, ~~ \text{if}~\left\|\mathbf{q}^{(k)}_{j}\right\|_{2} \!\geq \!\lambda\! / \!\mu^{(k)} \\
        			&0, ~~\text{otherwise;}
        		\end{aligned}\right.
        		\nonumber
        	\end{equation}
        	\vspace{-0.1cm}
			\STATE Update $\mathbf{Y}_{1}$, $\mathcal{Y}_{2}$, $\mathbf{Y}_{3}$, and $\mu$ by 
			\begin{equation}
        		\left\{\begin{array}{l}
        	\mathbf{Y}_{1}^{(k+1)}=\mathbf{Y}_{1}^{(k)}+\mu^{(k)}\left(\mathbf{X}-\mathbf{X}\mathbf{Z}^{(k+1)}-\mathbf{E}^{(k+1)}\right) \\
        	\mathcal{Y}_{2}^{(k\!+\!1)}\!(\!:,\!:,\!1\!)\!=\!\mathcal{Y}_{2}^{(k)}\!(\!:,\!:,\!1\!)\!+\!\mu^{(k)}\!\left(\mathbf{Z}^{(k\!+\!1)}\!-\!\mathcal{C}^{(k\!+\!1)}\!(\!:,\!:,\!1\!)\right)\\
        	\mathcal{Y}_{2}^{(k\!+\!1)}\!(\!:,\!:,\!2\!)\!=\!\mathcal{Y}_{2}^{(k)}\!(\!:,\!:,\!2\!)\!+\!\mu^{(k)}\!\left(\mathbf{B}^{(k\!+\!1)}\!-\!\mathcal{C}^{(k\!+\!1)}\!(\!:,\!:,\!2\!)\right) \\
    \mathbf{Y}_{3}^{(k+1)}=\mathbf{Y}_{3}^{(k)}+\mu^{(k)}\left(\mathbf{B}^{(k+1)}-\mathbf{D}^{(k+1)}\right) \\
        	\mu^{(k+1)}=\min \left(\rho \mu^{(k)}; \mu_{\max }\right);
        		\end{array}\right.
        		\nonumber 
        	\end{equation}
			\UNTIL convergence
		\end{algorithmic}  
	\end{algorithm}

\begin{figure*}[!t]
		\centering
		\subfloat[]{\includegraphics[width=3.8cm]{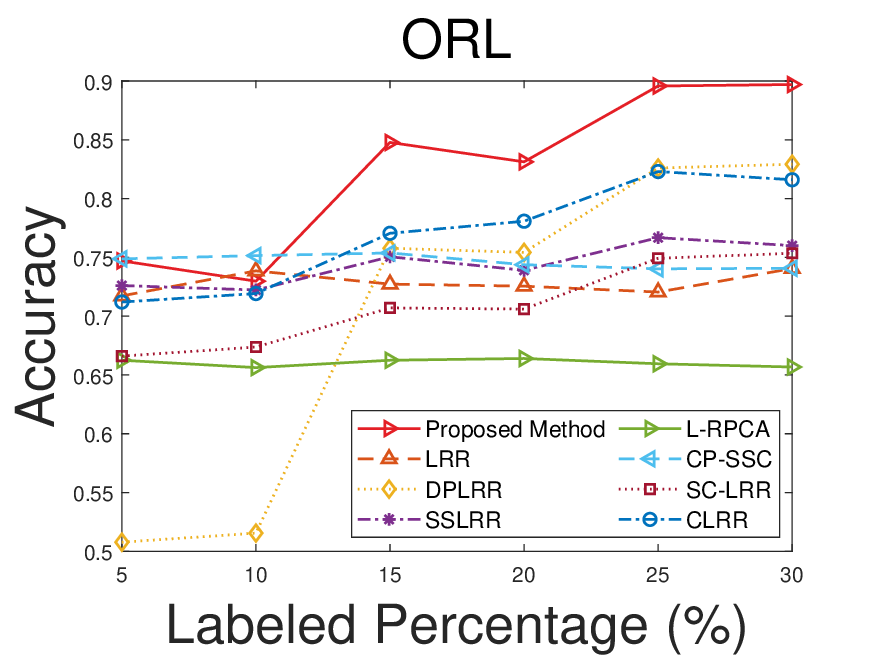}
			\label{fig_1_case}}
		\subfloat[]{\includegraphics[width=3.8cm]{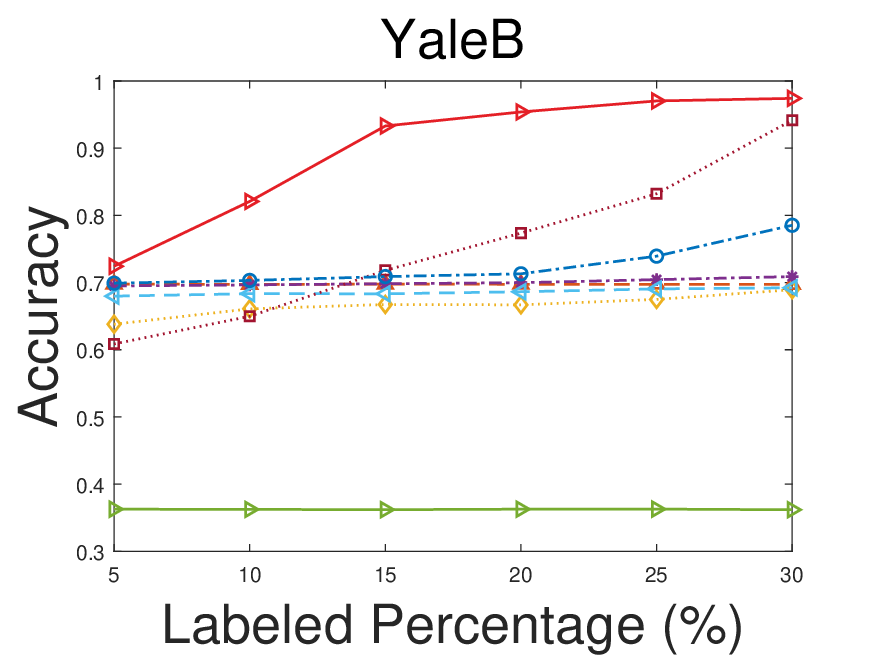}
			\label{fig_3_case}}
		\subfloat[]{\includegraphics[width=3.8cm]{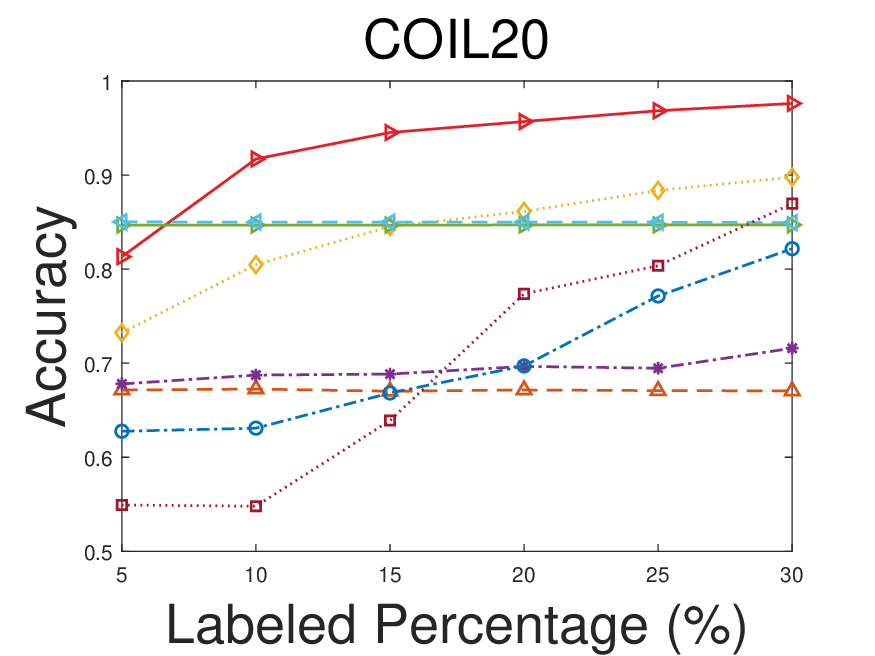}
			\label{fig_5_case}}
		\subfloat[]{\includegraphics[width=3.8cm]{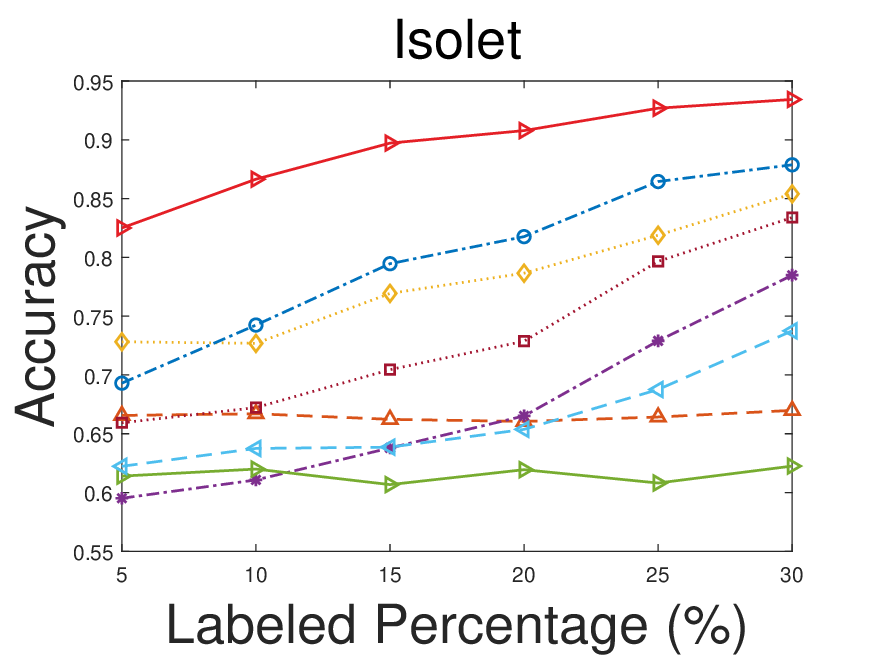}
			\label{fig_7_case}}
        \quad
		\subfloat[]{\includegraphics[width=3.8cm]{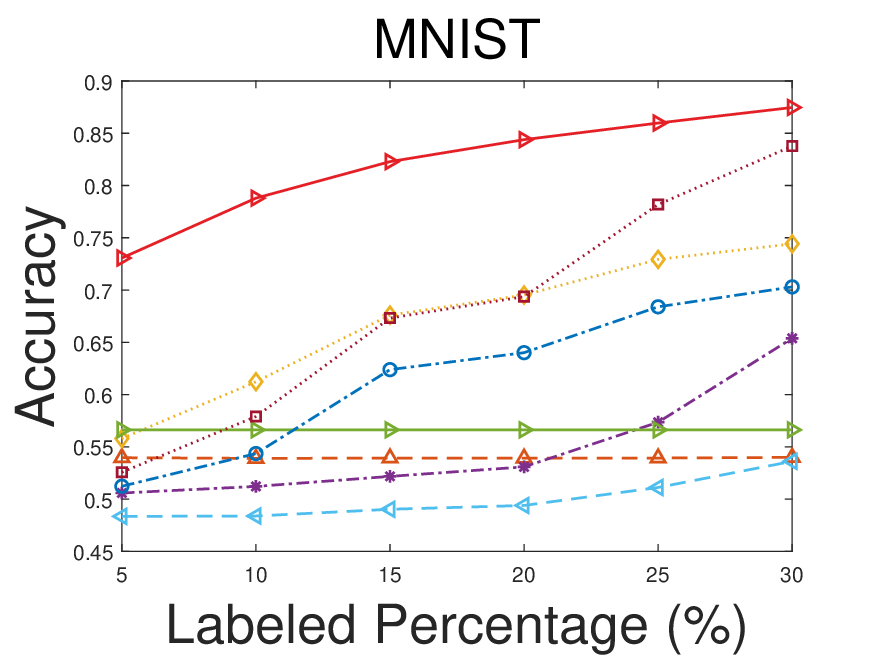}
			\label{fig_9_case}}
		\subfloat[]{\includegraphics[width=3.8cm]{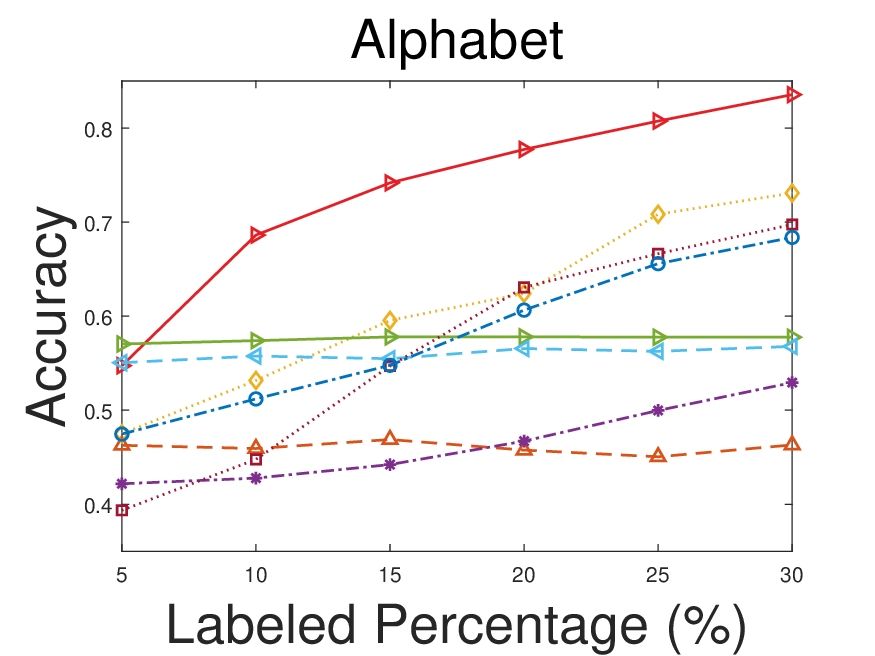}
			\label{fig_11_case}}
   		\subfloat[]{\includegraphics[width=3.8cm]{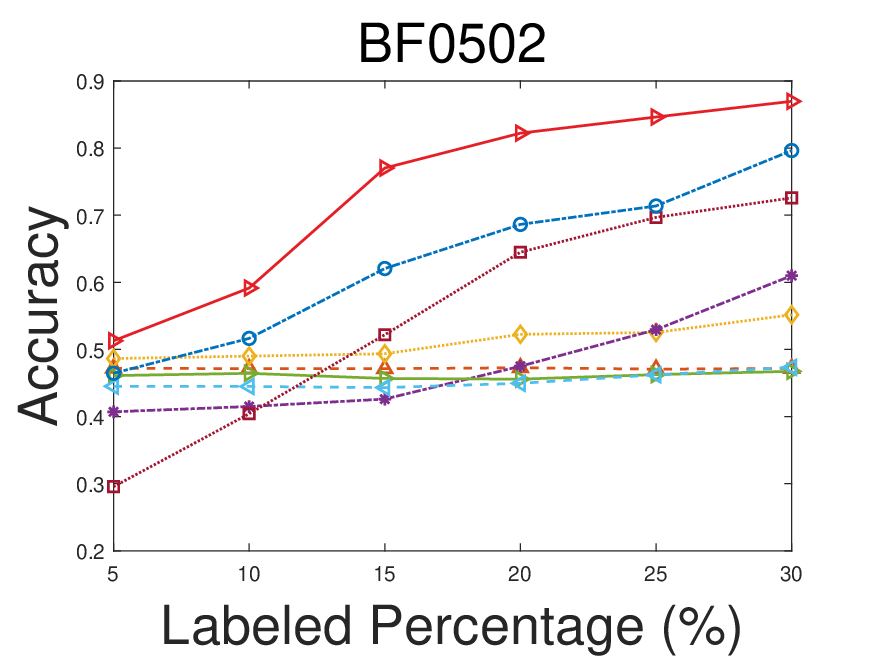}
			\label{fig_13_case}}
      \subfloat[]{\includegraphics[width=3.8cm]{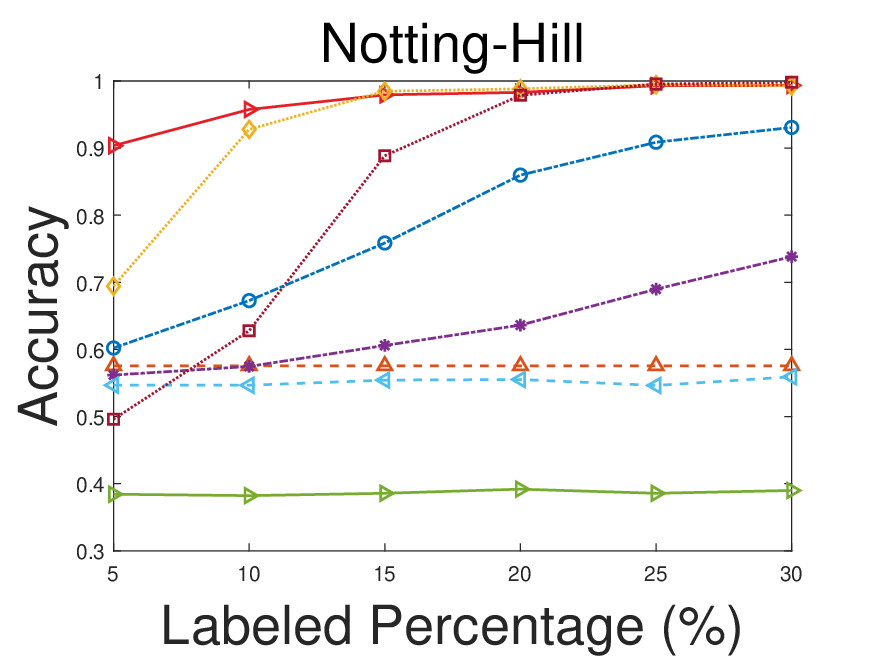}
		\label{fig_15_case}}
		\vspace{-0.2cm}
		\caption{\small{Comparisons of the accuracy of all methods under various supervisory information. All subfigures share the same legend.}}
		\label{fig_1}
	\end{figure*}
 
 	\begin{figure*}[!t]
        \vspace{-0.8cm}
		\centering
		\subfloat[]{\includegraphics[width=3.8cm]{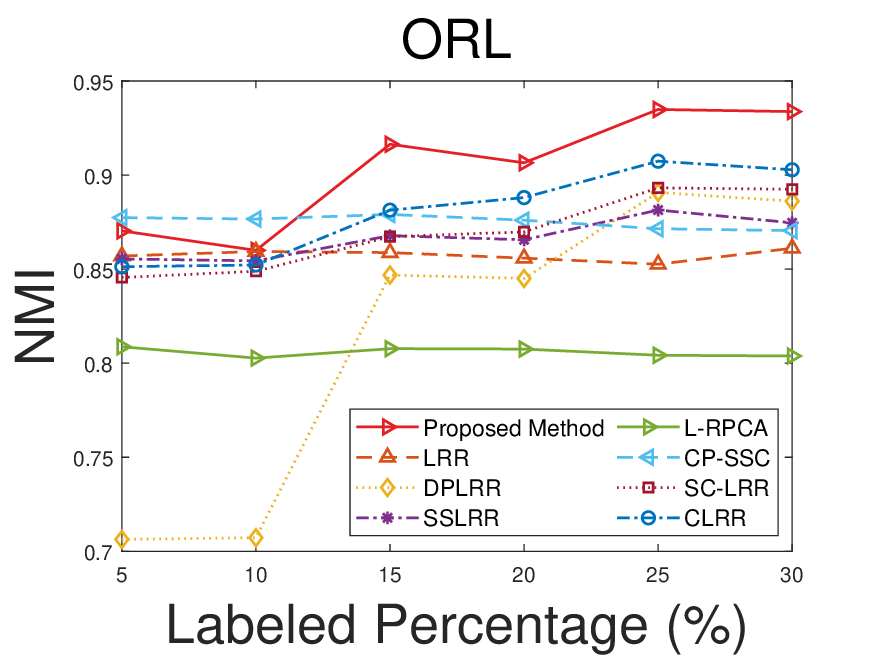}
			\label{fig_2_case}}
		\subfloat[]{\includegraphics[width=3.8cm]{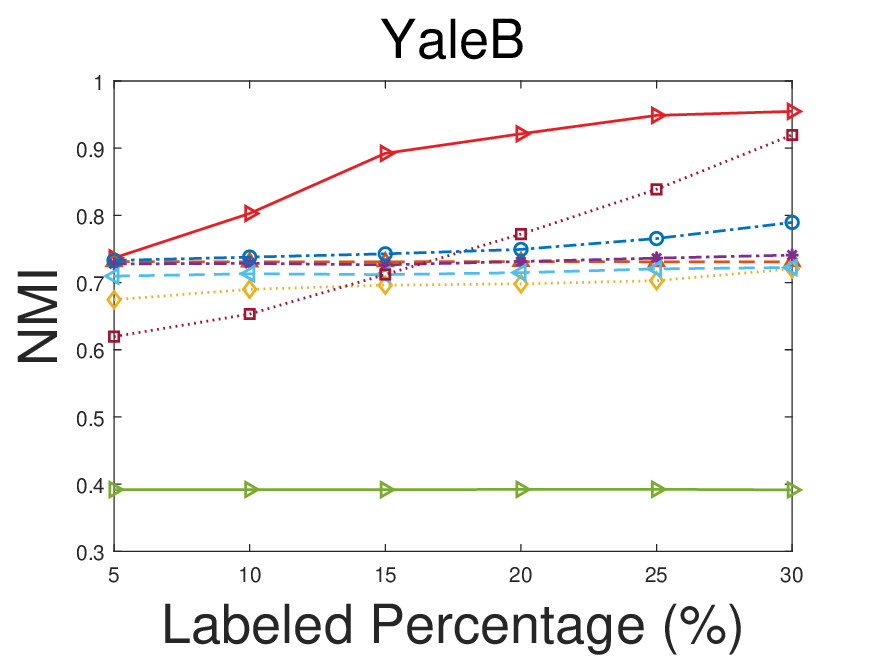}
			\label{fig_4_case}}
		\subfloat[]{\includegraphics[width=3.8cm]{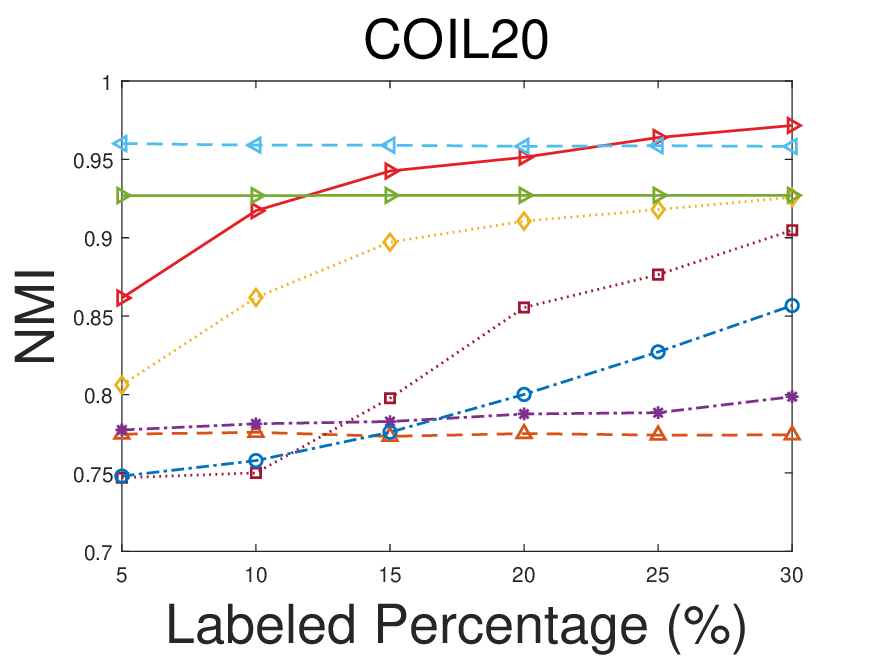}
			\label{fig_6_case}}
		\subfloat[]{\includegraphics[width=3.8cm]{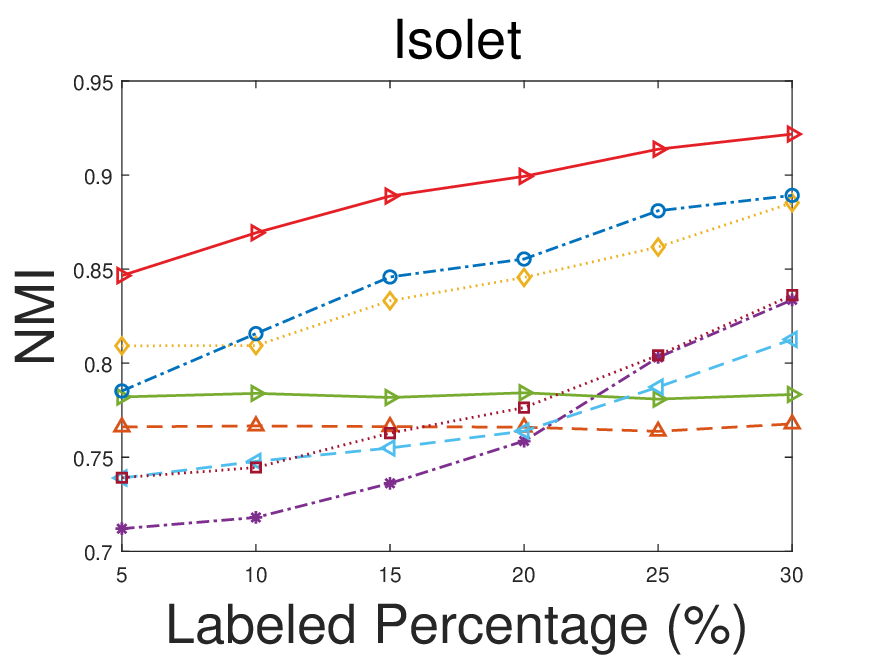}
			\label{fig_8_case}}
        \quad
		\subfloat[]{\includegraphics[width=3.8cm]{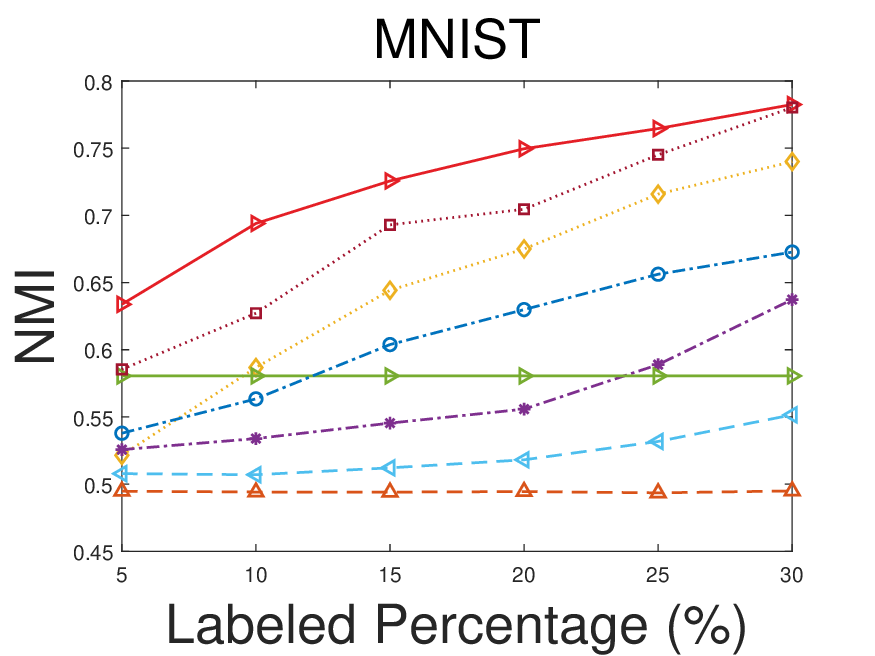}
			\label{fig_10_case}}
		\subfloat[]{\includegraphics[width=3.8cm]{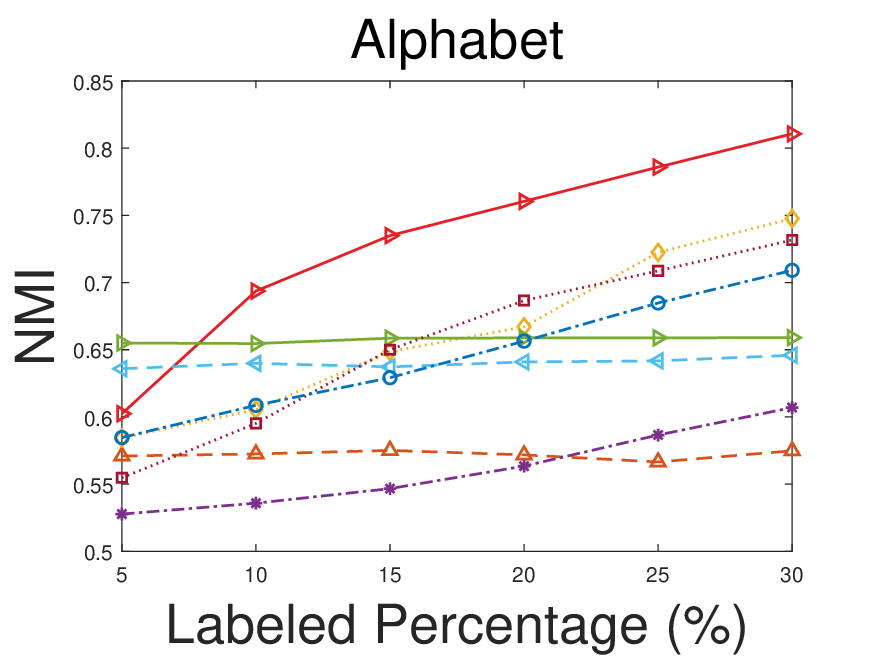}
			\label{fig_12_case}}
   	\subfloat[]{\includegraphics[width=3.8cm]{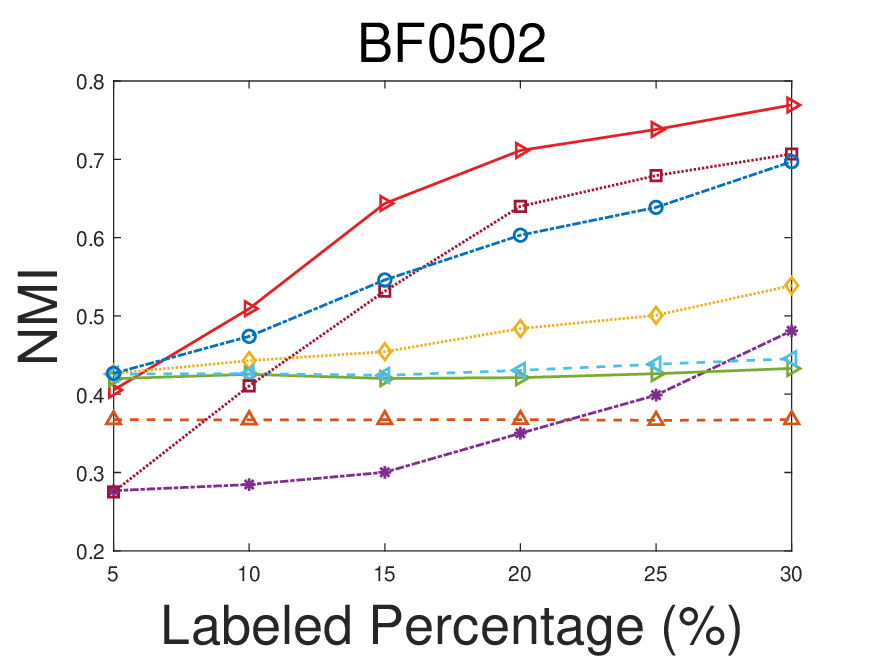}
			\label{fig_14_case}}
      \subfloat[]{\includegraphics[width=3.8cm]{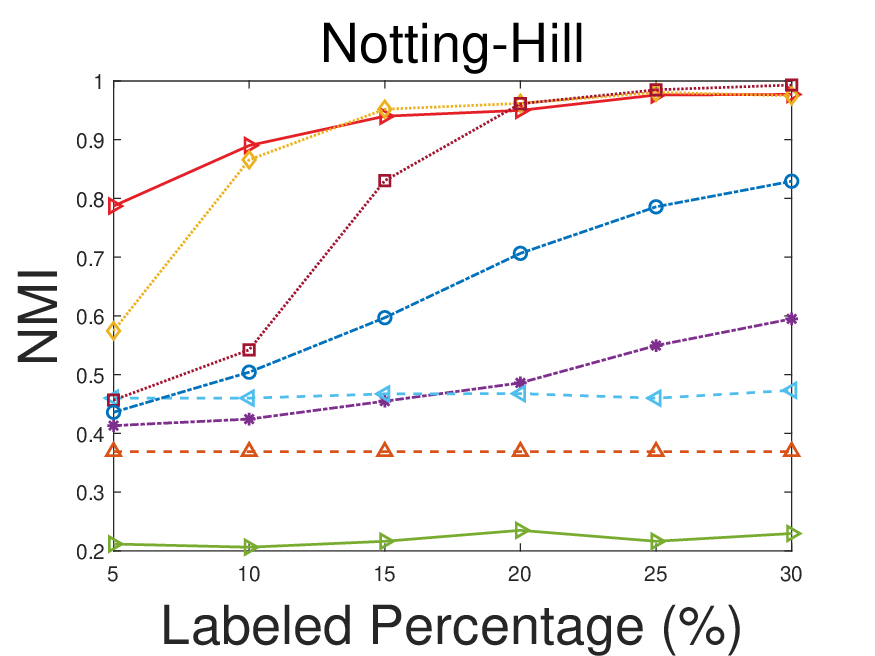}
		\label{fig_16_case}}
		\vspace{-0.2cm}
		\caption{\small{Comparisons of the NMI of all methods under various supervisory information. All subfigures share the same legend.}}
		\label{fig_2}
		\vspace{-0.2cm}
	\end{figure*}

 \begin{table*}[!t]
		\renewcommand{\arraystretch}{0.95}
		\setlength\tabcolsep{9pt}
		\caption{Detailed Comparisions of Accuracy and NMI under 30\% Initial Labels.}
		\label{table_1}
		\centering
		\begin{tabular}{cccccccccc}
			\hline
		 \textbf{Accuracy} & ORL & YaleB & COIL20 & Isolet & MNIST & Alphabet & BF0502 & Notting-Hill & \textbf{Average}\\
			\hline
                LRR & 0.7405 & 0.6974 & 0.6706 & 0.6699 & 0.5399 & 0.4631 & 0.4717 & 0.5756 & 0.6036 \\
                DPLRR & 0.8292 & 0.6894 & 0.8978 & 0.8540 & 0.7442 & 0.7309 & 0.5516 & 0.9928 & 0.7862 \\
                SSLRR & 0.7600 & 0.7089 & 0.7159 & 0.7848 & 0.6538 & 0.5294 & 0.6100 & 0.7383 & 0.6876 \\
                L-RPCA & 0.6568 & 0.3619 & 0.8470 & 0.6225 & 0.5662 & 0.5776 & 0.4674 & 0.3899 & 0.5612 \\
                CP-SSC & 0.7408 & 0.6922 & 0.8494 & 0.7375 & 0.5361 & 0.5679 & 0.4733 & 0.5592 & 0.6445 \\
                SC-LRR & 0.7535 & 0.9416 & 0.8696 & 0.8339 & 0.8377 & 0.6974 & 0.7259 & \textbf{0.9982} & 0.8322 \\
				CLRR & 0.8160 & 0.7853 & 0.8217 & 0.8787 & 0.7030 & 0.6837 & 0.7964 & 0.9308 & 0.8020 \\
			Proposed Method & \textbf{0.8965} & \textbf{0.9742} & \textbf{0.9761} & \textbf{0.9344} & \textbf{0.8747} & \textbf{0.8355} & \textbf{0.8697} & 0.9934 & \textbf{0.9193} \\ 
			\hline
       		\textbf{NMI} & ORL & YaleB & COIL20 & Isolet & MNIST & Alphabet & BF0502 & Notting-Hill & \textbf{Average}\\
			\hline
                LRR & 0.8611 & 0.7309 & 0.7742 & 0.7677 & 0.4949 & 0.5748 & 0.3675 & 0.3689 & 0.6175 \\
                DPLRR & 0.8861 & 0.7205 & 0.9258 & 0.8853 & 0.7400 & 0.7477 & 0.5388 & 0.9748 & 0.8024 \\
			    SSLRR & 0.8746 & 0.7409 & 0.7986 & 0.8337 & 0.6373 & 0.6070 & 0.4810 & 0.5949 & 0.6960 \\
                L-RPCA & 0.8038 & 0.3914 & 0.9271 & 0.7834 & 0.5805 & 0.6590 & 0.4329 & 0.2294 & 0.6009 \\
                 CP-SSC & 0.8705 & 0.7224 & 0.9583 & 0.8127 & 0.5516 & 0.6459 & 0.4453 & 0.4733 & 0.6850 \\
                 SC-LRR & 0.8924 & 0.9197 & 0.9048 & 0.8362 & 0.7803 & 0.7316 & 0.7068 & \textbf{0.9931} & 0.8456 \\
                CLRR & 0.9028 & 0.7895 & 0.8568 & 0.8892 & 0.6727 & 0.7091 & 0.6970 & 0.8293 & 0.7933 \\
			Proposed Method & \textbf{0.9337} & \textbf{0.9548} & \textbf{0.9716} & \textbf{0.9218} & \textbf{0.7825} & \textbf{0.8107} & \textbf{0.7693} & 0.9771 & \textbf{0.8902} \\
			\hline
		\end{tabular}
  \end{table*}

  \begin{table*}[!t]
		\renewcommand{\arraystretch}{0.95}
		\setlength\tabcolsep{8pt}
		\caption{Ablation Study.}
		\label{table_ablation}
		\centering
		\begin{tabular}{ccccccccccc}
			\hline
			Percentage & \textbf{Accuracy} & ORL & YaleB & COIL20 & Isolet & MNIST & Alphabet & BF0502 & Notting-Hill & \textbf{Average}\\ 
			\hline
			\multirow{5}{*}{10} 
			& SSLRR & 0.7223 & 0.6965 & 0.6874 & 0.6107 & 0.5121 & 0.4278 & 0.4150 & 0.5747 & 0.5808 \\
			& CLRR & 0.7193 & 0.7032 & 0.6309 & 0.7424 & 0.5435 & 0.5120 & 0.5165 & 0.6728 & 0.6301 \\
			& Eq. \eqref{eq_origin_woL} & 0.7298 & 0.7838 & 0.6744 & 0.8599 & 0.5224 & 0.5022 & 0.5786 & 0.8079 & 0.6824 \\
			& Eq. \eqref{eq_origin} & 0.7298 & 0.7838 & 0.8708 & 0.8424 & 0.7659 & 0.6640 & 0.5779 & 0.9573 & 0.7740 \\
			& Eq.  \eqref{eqn_repair} & \textbf{0.7523} & \textbf{0.8696} & \textbf{0.9171} & \textbf{0.8665} & \textbf{0.7879} & \textbf{0.6862} & \textbf{0.5915} & \textbf{0.9576} & \textbf{0.8036}\\
			
			\hline
			\multirow{5}{*}{20}
			& SSLRR & 0.7390 & 0.6998 & 0.6966 & 0.6651 & 0.5308 & 0.4672 & 0.4750 & 0.6363 & 0.6137 \\
			& CLRR & 0.7808 & 0.7130 & 0.6971 & 0.8176 & 0.6401 & 0.6064 & 0.6863 & 0.8598 & 0.7251 \\
			& Eq. \eqref{eq_origin_woL} & 0.7860 & 0.9194 & 0.8101 & 0.9012 & 0.6661 & 0.6443 & 0.7554 & 0.9378 & 0.8025 \\
			& Eq. \eqref{eq_origin} & 0.7860 & 0.9194 & 0.9364 & 0.9065 & 0.8366 & 0.7511 & 0.8077 & 0.9817 & 0.8657 \\
			& Eq. \eqref{eqn_repair} & \textbf{0.8325} & \textbf{0.9548} & \textbf{0.9569} & \textbf{0.9078} & \textbf{0.8439} & \textbf{0.7772} & \textbf{0.8223} & \textbf{0.9831} & \textbf{0.8848}
			\\
			
			\hline
			\multirow{5}{*}{30} 
			& SSLRR & 0.7600 & 0.7089 & 0.7159 & 0.7848 & 0.6538 & 0.5294 & 0.6100 & 0.7383 & 0.6876 \\
			& CLRR & 0.8160 & 0.7853 & 0.8217 & 0.8787 & 0.7030 & 0.6837 & 0.7964 & 0.9308 & 0.8020 \\
			& Eq. \eqref{eq_origin_woL} & 0.8893 & 0.9664 & 0.9096 & 0.9222 & 0.8370 & 0.7671 & 0.8083 & 0.9661 & 0.8832 \\
			& Eq. \eqref{eq_origin} & 0.8893 & 0.9664 & 0.9710 & 0.9300 & 0.8745 & 0.8244 & 0.8631 & 0.9917 & 0.9138 \\
			& Eq. \eqref{eqn_repair} & \textbf{0.8965} & \textbf{0.9742} & \textbf{0.9761} & \textbf{0.9344} & \textbf{0.8747} & \textbf{0.8355} & \textbf{0.8697} & \textbf{0.9934} & \textbf{0.9193} \\ 
			\hline
		\end{tabular}
	\end{table*}

	\section{Proposed Method}\label{section_3}
	\subsection{Model Formulation}
 
	As aforementioned, existing semi-supervised subspace clustering methods usually impose the pairwise constraints on the affinity matrix in a simple element-wise manner, which under-uses the supervisory information to some extent. 
	As studied in previous works \cite{6180173, 9454509}, the ideal affinity matrix $\mathbf{Z}$ is low-rank as a sample should be only reconstructed by the samples within the same class.
	Meanwhile, the ideal pairwise constraint matrix $\mathbf{B}$ is also low-rank, as it records the pairwise relationship among samples.
	Moreover, we observe that their low-rank structures should be identical. 
	Accordingly, if 
	we stack them to form a 3-D tensor $\mathcal{C}\in\mathbb{R}^{n\times n\times 2}$, i.e., $\mathcal{C}(:,:,1)=\mathbf{Z}$, and $\mathcal{C}(:,:,2)=\mathbf{B}$, the formed tensor $\mathcal{C}$ should be low-rank ideally.
	Therefore, we use a global tensor low-rank norm to exploit this prior and preliminarily formulate the problem as
	\begin{equation}
		\begin{aligned}
			\min _{\mathcal{C}, \mathbf{E}, \mathbf{B}, \mathbf{Z} }&\quad\|\mathcal{C}\|_{\circledast}+\lambda\|\mathbf{E}\|_{2,1} \\
			\text {s.t.}&\quad\mathbf{X}=\mathbf{X Z}+\mathbf{E}, \mathcal{C}(:,:,1)=\mathbf{Z}, \mathcal{C}(:,:,2)=\mathbf{B},  \\
			&\quad\mathbf{B}_{i j}=s,(i, j) \in \mathbf{\Omega}_{m}, \mathbf{B}_{i j}=-s,(i, j) \in \mathbf{\Omega}_{c}. 
		\end{aligned}
		\label{eq_origin_woL}
	\end{equation}
	In Eq. \eqref{eq_origin_woL}, we adopt the nuclear norm $\|\cdot\|_{\circledast}$ defined on tensor-SVD \cite{8606166} to seek the low-rank representation, and other kinds of tensor low-rank norms are also applicable, e.g., \cite{9763010}.
	We introduce a scalar $s$ to constrain the maximum and minimum values of $\mathbf{B}$, promoting that $\mathbf{B}$ has a similar scale to $\mathbf{Z}$. 
	Empirically, $s$ is set to the largest element of the learned affinity by LRR.
	By solving Eq. \eqref{eq_origin_woL}, the affinity matrix $\mathbf{Z}$ and pairwise constraint matrix $\mathbf{B}$ are jointly optimized according to the nuclear norm on $\mathcal{C}$, i.e., the supervisory information is transferred to $\mathbf{Z}$, and at the same time, the learned affinity matrix can also augment the initial pairwise constraints from a global perspective.
 
	Besides, if two samples ($\mathbf{x}_i$ and $\mathbf{x}_j$) are close to each other in the feature space, we can expect they have a similar pairwise relationship, i.e., $\mathbf{B}(:,i)$ is close to $\mathbf{B}(:,j)$. 
	To encode this prior, we first construct a $k$NN graph $\mathbf{W}\!\in\!\mathbb{R}^{n\times n}$ to capture the local geometric structure of samples, and use the local Laplacian regularization $\operatorname{Tr}(\mathbf{B L B^{\top}})$ to replenish the global low-rank term, where $\mathbf{L}\!=\!\mathbf{D}\!-\!\mathbf{W}$ is the Laplacian matrix with $\mathbf{D}_{ii}\!=\!\sum_{j} \mathbf{W}_{ij}$ \cite{chen2020multi}.
	Therefore, our model is finally formulated as 
	\vspace{-0.2cm}
	\begin{equation}
		\begin{aligned}
			\min _{\mathcal{C}, \mathbf{E}, \mathbf{B}, \mathbf{Z} }&\quad\|\mathcal{C}\|_{\circledast}+\lambda\|\mathbf{E}\|_{2,1}+\beta\operatorname{Tr}(\mathbf{B L B}^{\top}) \\
			\text{s.t.}&\quad\mathbf{X}=\mathbf{X Z}+\mathbf{E}, \mathcal{C}(:,:,1)=\mathbf{Z}, \mathcal{C}(:,:,2)=\mathbf{B},  \\
			&\quad\mathbf{B}_{i j}=s,(i, j) \in \mathbf{\Omega}_{m}, \mathbf{B}_{i j}=-s,(i, j) \in \mathbf{\Omega}_{c}. 
		\end{aligned}
		\label{eq_origin}
	\end{equation}
	
	After solving Eq. \eqref{eq_origin}, we first normalize each column of $\mathbf{Z}$ to $\left[0,1\right]$, and normalize $\mathbf{B}$ by $\mathbf{B}\leftarrow\mathbf{B}/s$. Then, we use the augmented pairwise constraint matrix $\mathbf{B}$ to repair $\mathbf{Z}$, i.e.,  
	\begin{equation}
		\mathbf{Z}_{ij}\leftarrow\left\{\begin{aligned}
			&1-\left(1-\mathbf{B}_{i j}\right)\left(1-\mathbf{Z}_{i j}\right), ~~\text{if}~ \mathbf{B}_{i j} \geq 0 \\ 
			&\left(1+\mathbf{B}_{i j}\right) \mathbf{Z}_{i j}, ~~\text{if}~ \mathbf{B}_{i j}<0.
		\end{aligned}\right.
		\label{eqn_repair}
	\end{equation}
	When $\mathbf{B}_{i j}$ is larger than 0, $\mathbf{x}_i$ and $\mathbf{x}_j$ are likely to belong to the same class, Eq. \eqref{eqn_repair} will increase the corresponding element of $\mathbf{Z}$. 
	Similarly, when $\mathbf{B}_{i j}$ is less than 0, $\mathbf{Z}_{i j}$ will be depressed.
	Therefore, Eq. \eqref{eqn_repair} further enhances the affinity matrix by the augmented pairwise constraints.
	Finally, we apply spectral clustering \cite{8587179} on $\mathbf{W}\!=\!(|\mathbf{Z}|\!+\!|\mathbf{Z}^{\!\top}|)\!/2$ to get the subspace segmentation.
	
    \subsection{Optimization Algorithm}
    As Eq. \eqref{eq_origin} contains multiple variables and constraints, we solve it by alternating direction method of multipliers (ADMM) \cite{8962252}.
	Algorithm \ref{alg1} summarizes the whole pseudo code. Due to the page limitation, the detailed derivation process can be found in the supplementary file. 
	
	The computational complexity of Algorithm \ref{alg1} is dominated by steps 3-5.  
 Specifically, the step 3 solves the t-SVD of an $n\times n \times 2$ tensor with the complexity of $\mathcal{O}(2n^{2}\log 2 + 2n^3)$ \cite{8606166}. 
	The steps 4-5 involve matrix inverse and matrix multiplication operations with the complexity of $\mathcal{O}(n^{3})$. 
	Note that in step 4, the to be inversed matrix $\left(\mathbf{I}+\mathbf{X}^{\top} \mathbf{X}\right)$ is unchanged, which only needs to be calculated once in advanced.
	In summary, the overall computational complexity of Algorithm \ref{alg1} is $\mathcal{O}(n^3)$ in one iteration.

\section{Experiments}\label{section_4}
	\begin{figure}[!t]
        \vspace{-0.5cm}
		\centering
		\subfloat[Proposed]{\includegraphics[width=2.2cm]{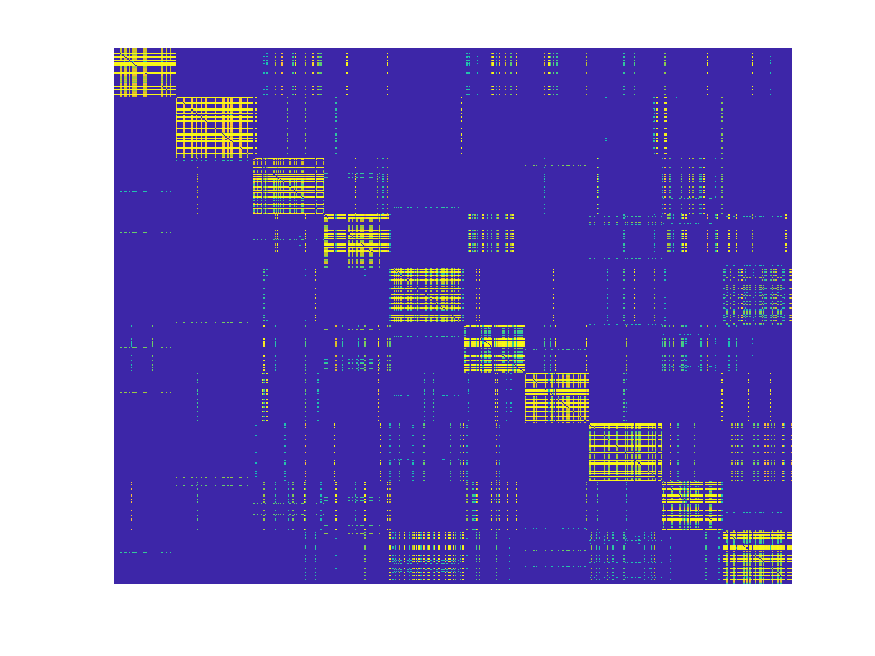}
			\label{fig_tnn}}
		\subfloat[LRR]{\includegraphics[width=2.2cm]{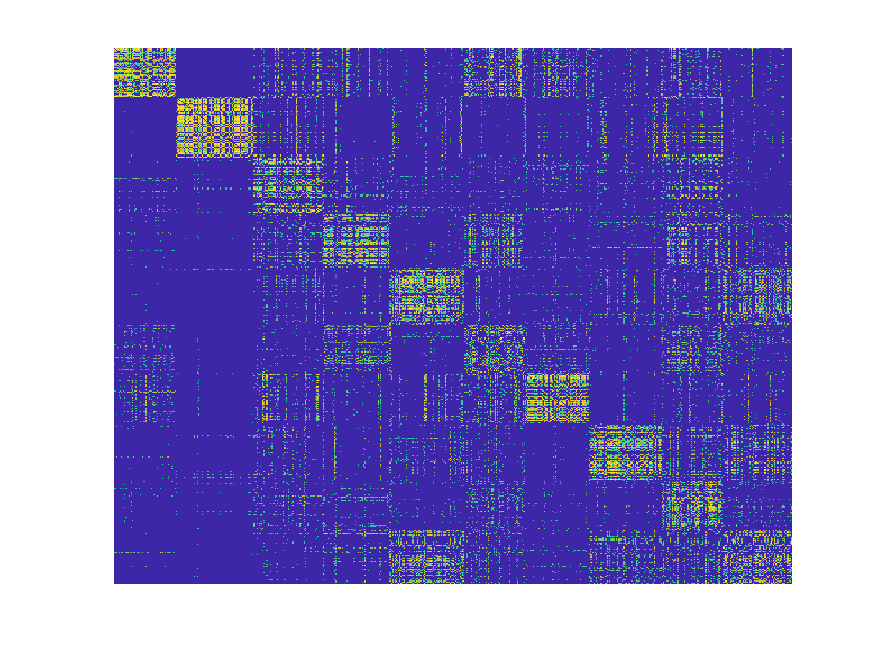}
			\label{fig_lrr}}
		\subfloat[DPLRR]{\includegraphics[width=2.2cm]{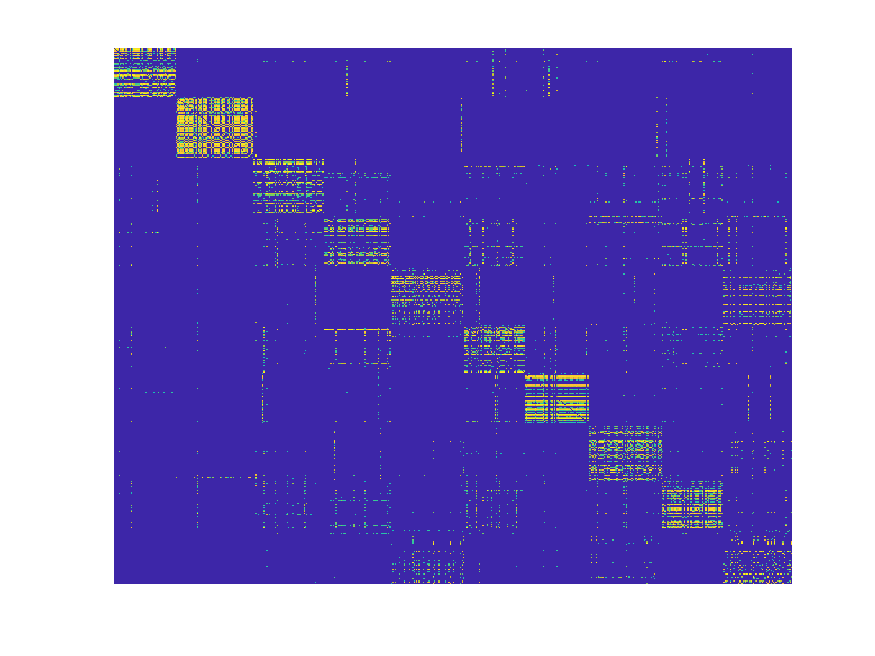}
			\label{fig_dplrr}}
		\subfloat[SSLRR]{\includegraphics[width=2.2cm]{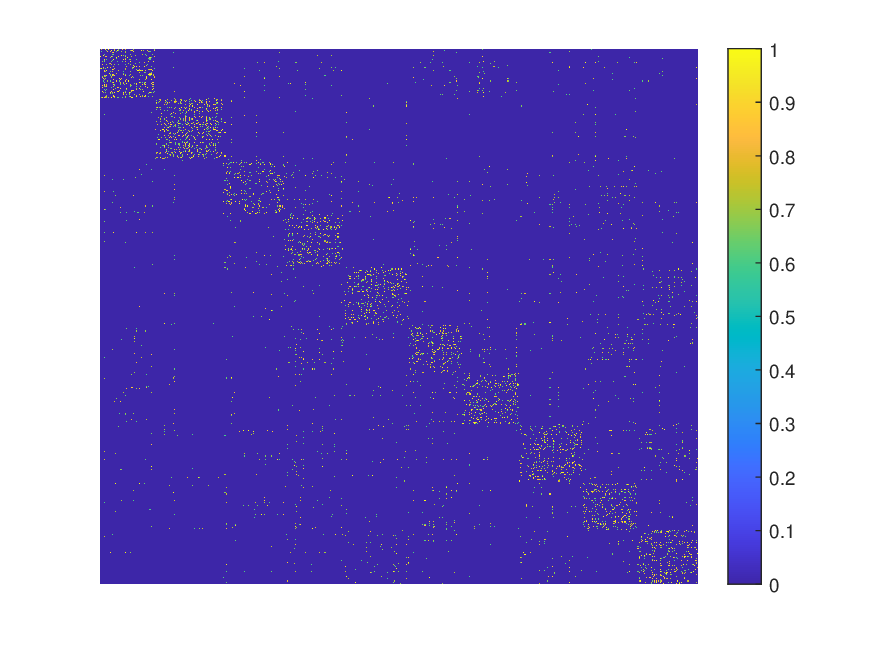}
			\label{fig_sslrr}}
		\quad
		\subfloat[L-RPCA]{\includegraphics[width=2.2cm]{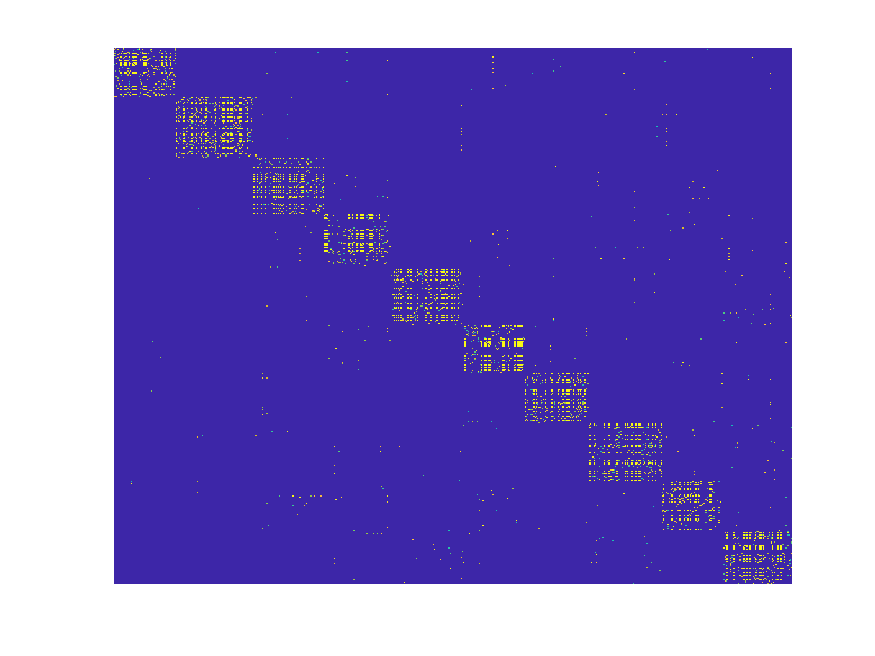}
			\label{fig_lrpca}}
		\subfloat[CP-SSC]{\includegraphics[width=2.2cm]{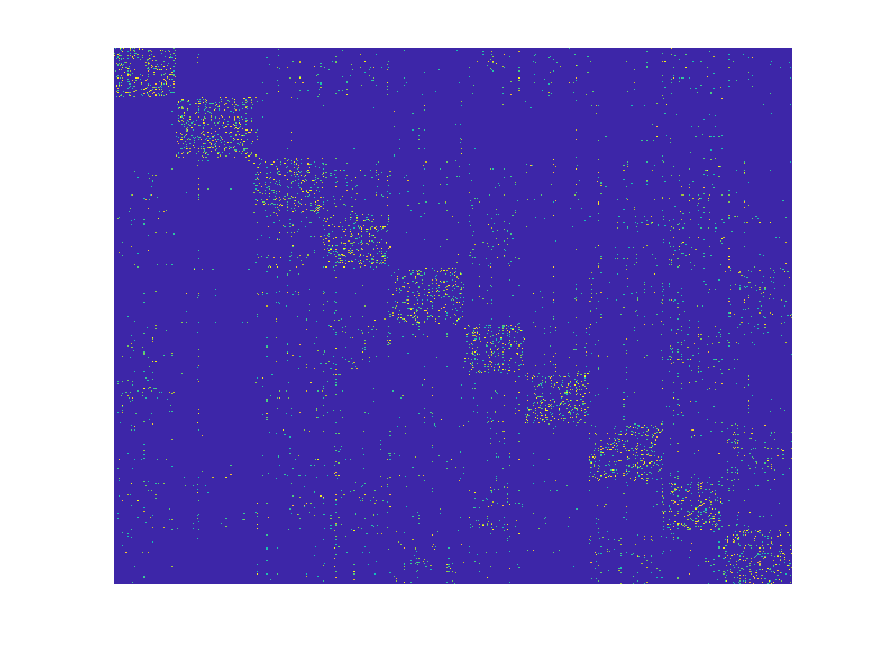}
			\label{fig_cpssc}}
		\subfloat[SC-LRR]{\includegraphics[width=2.2cm]{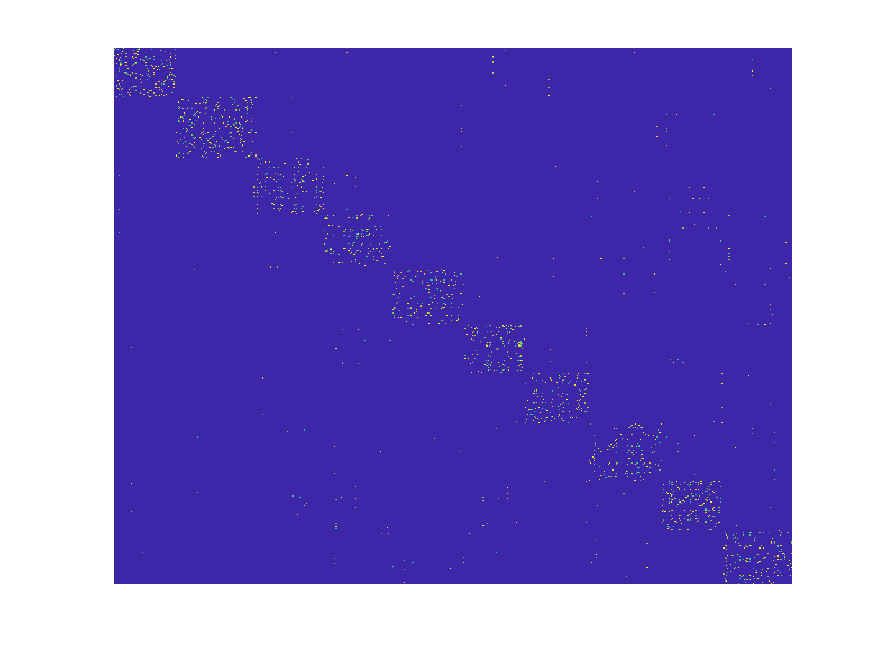}
			\label{fig_sclrr}}
		\subfloat[CLRR]{\includegraphics[width=2.2cm]{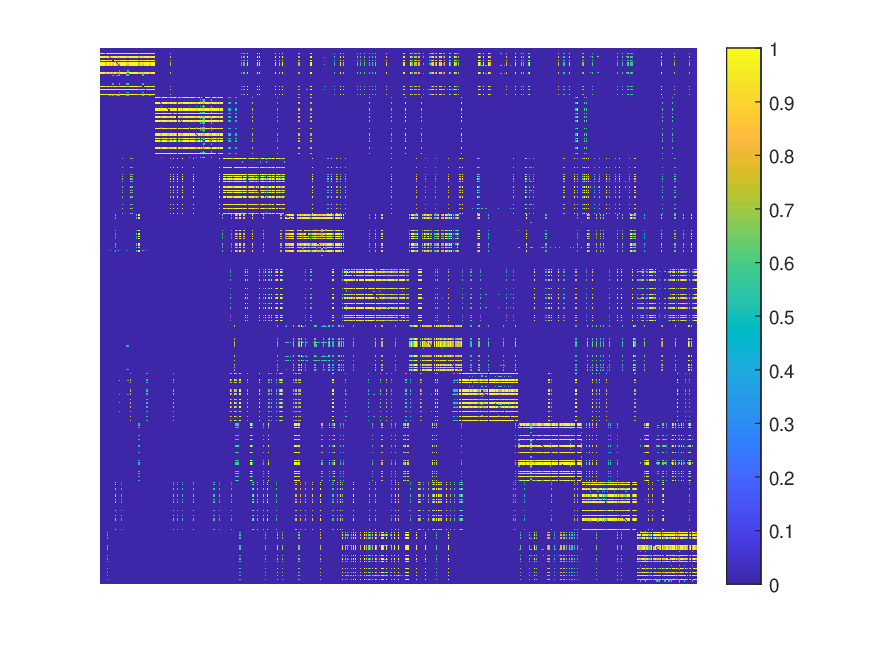}
			\label{fig_clrr}}
		\caption{Visual comparison of the affinity matrices learned by different methods on MNIST. The learned affinity matrices were normalized to [0,1]. Zoom in the figure for a better view.}
		\label{fig_visual}
		\vspace{-0.5cm}
	\end{figure}
 
	\begin{figure}[!t]
		\centering
		\subfloat[]{\includegraphics[width=3cm]{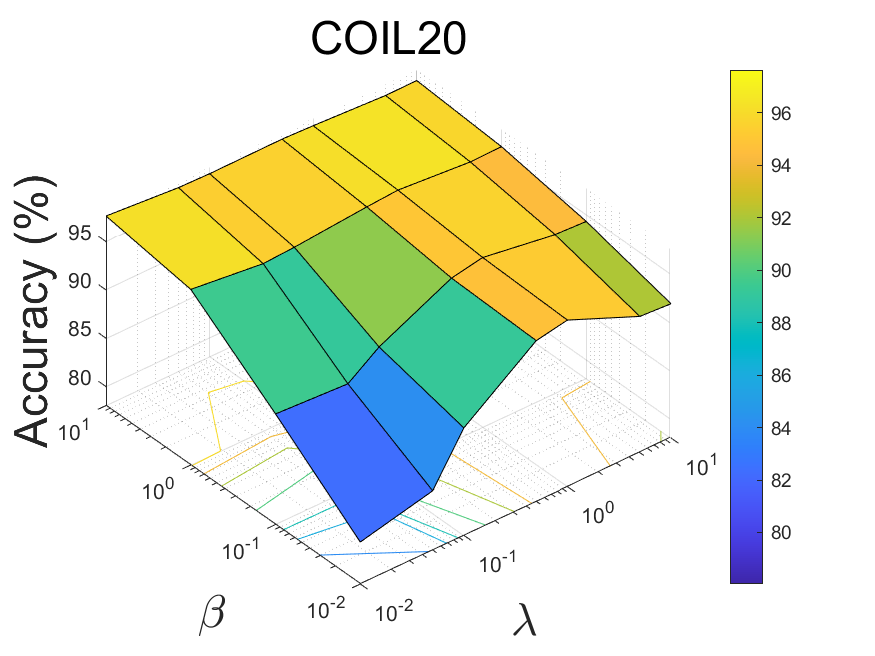}}
		\subfloat[]{\includegraphics[width=3cm]{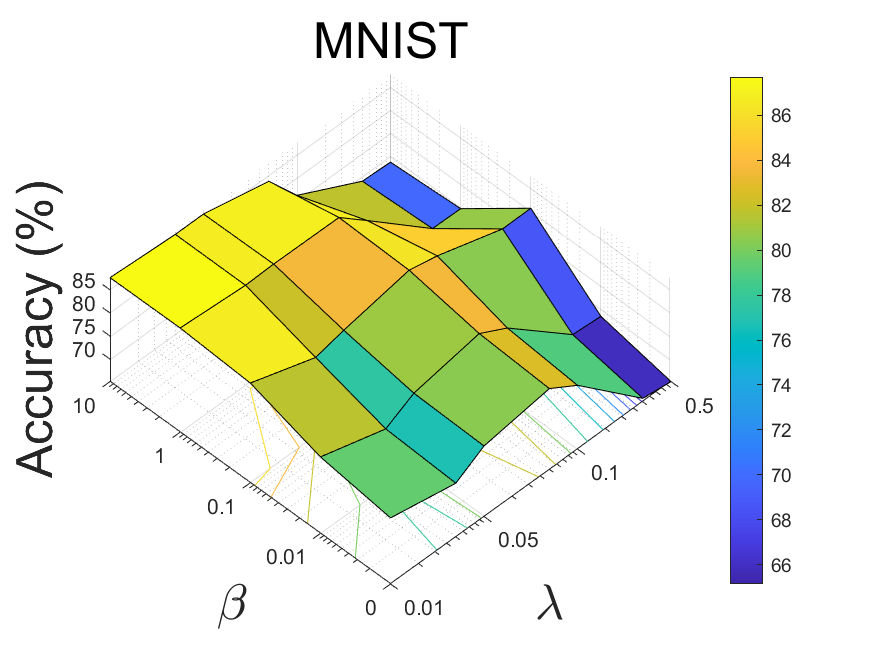}}
		\subfloat[]{\includegraphics[width=3cm]{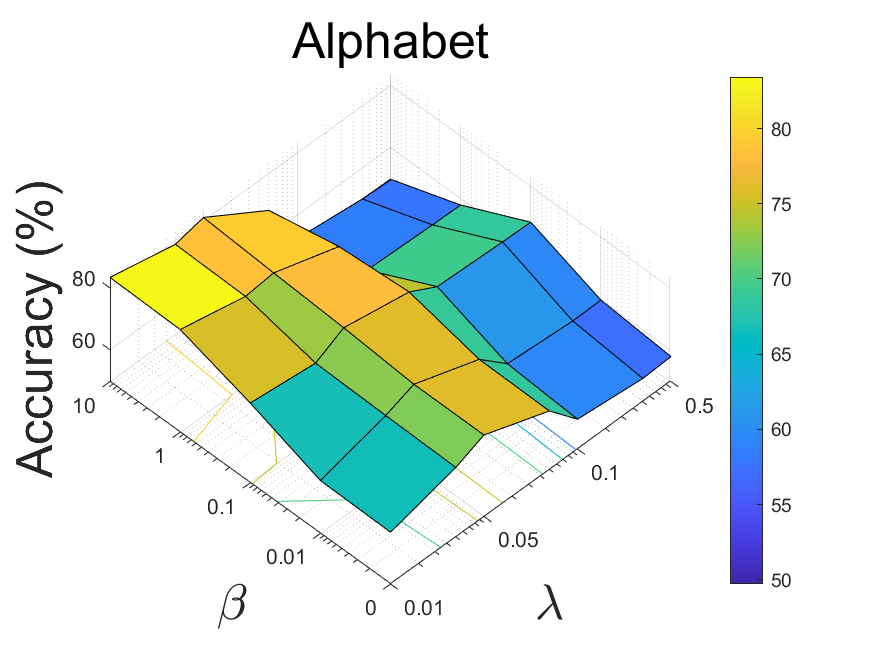}}
		\caption{Influence of the hyper-parameters on clustering performance.}
		\vspace{-0.5cm}
		\label{fig_3}
	\end{figure}

 \begin{figure*}[htbp]
            \vspace{-0.3cm}
            \centering
            \subfloat[]{\includegraphics[width=3.8cm]{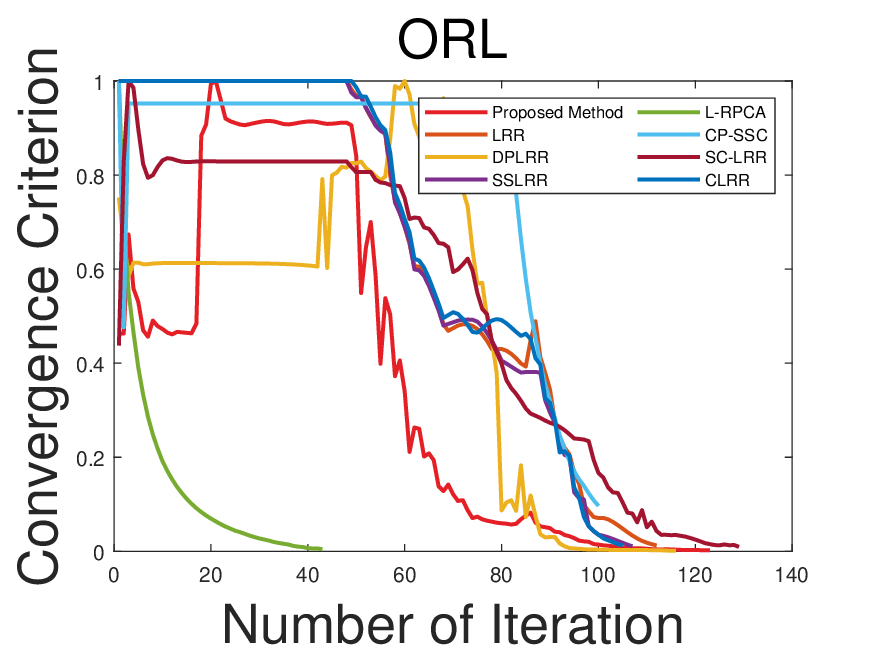}
                \label{fig_1_converge}}
            \subfloat[]{\includegraphics[width=3.8cm]{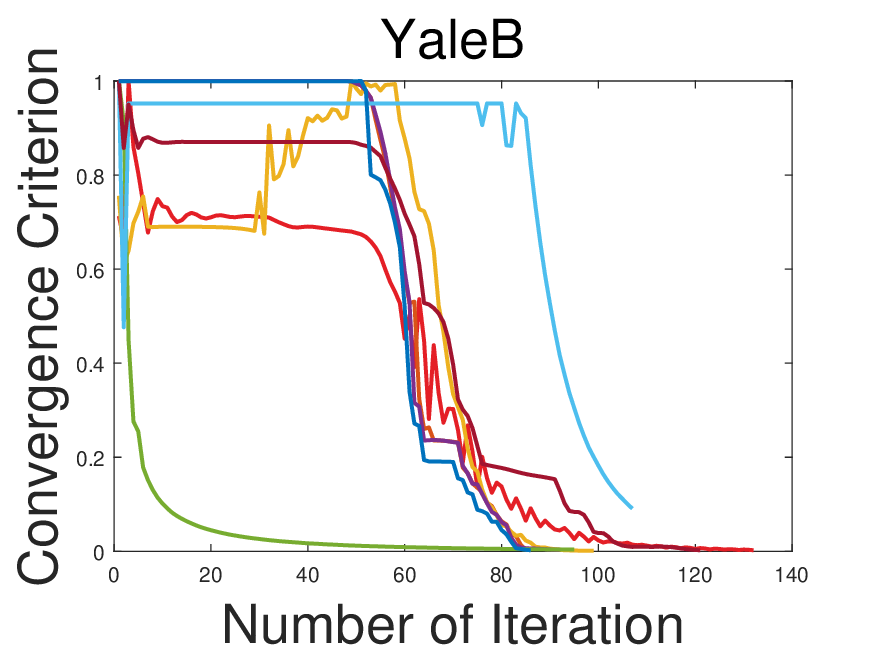}
                \label{fig_3_converge}}
            \subfloat[]{\includegraphics[width=3.8cm]{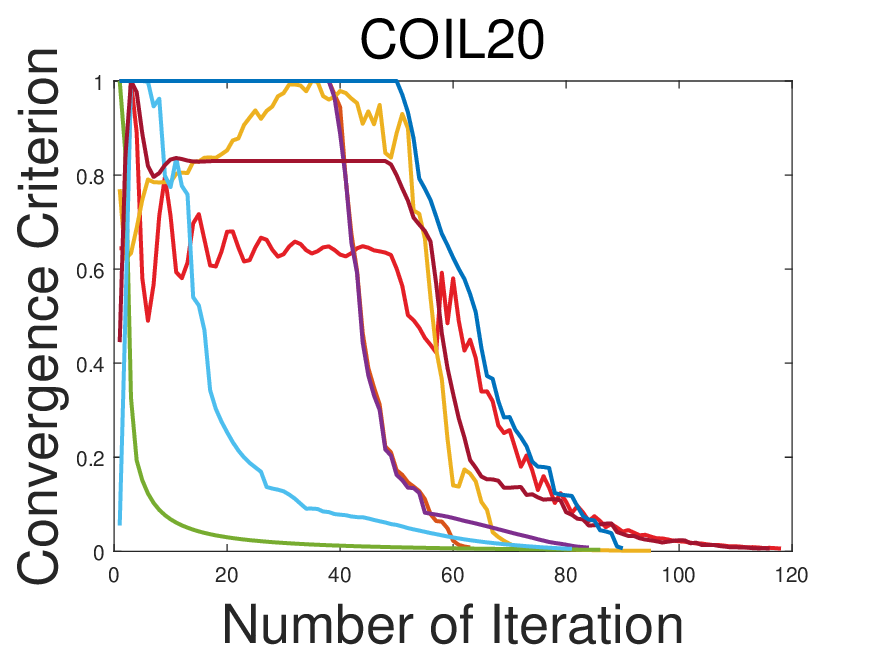}
                \label{fig_5_converge}}
            \subfloat[]{\includegraphics[width=3.8cm]{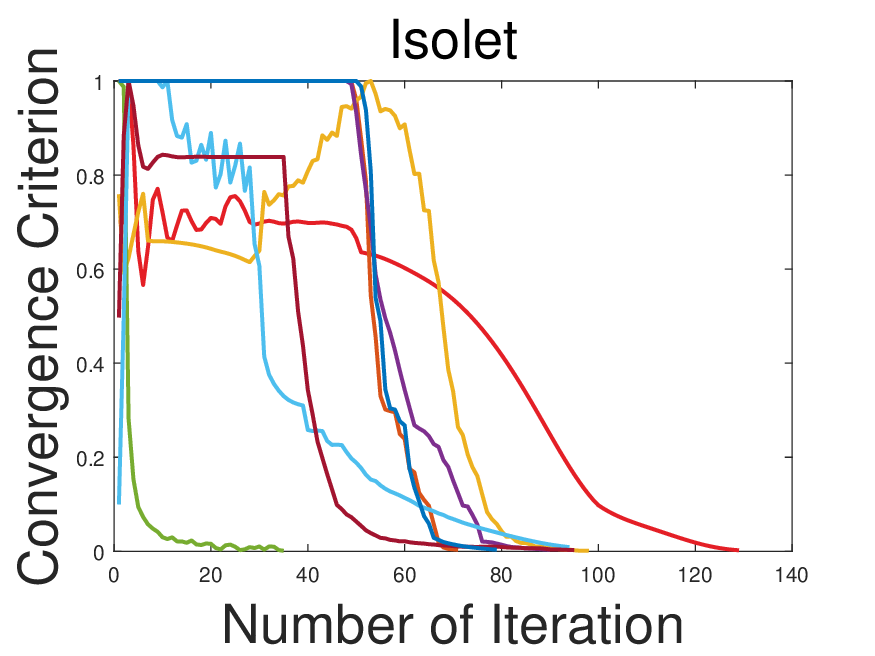}
                \label{fig_7_converge}}
            \quad
            \subfloat[]{\includegraphics[width=3.8cm]{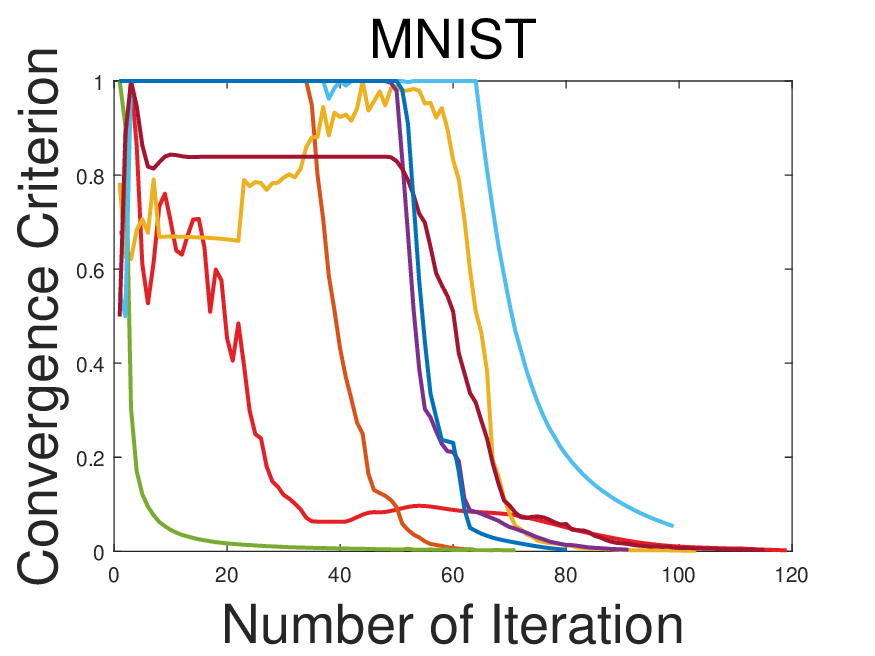}
                \label{fig_9_converge}}
            \subfloat[]{\includegraphics[width=3.8cm]{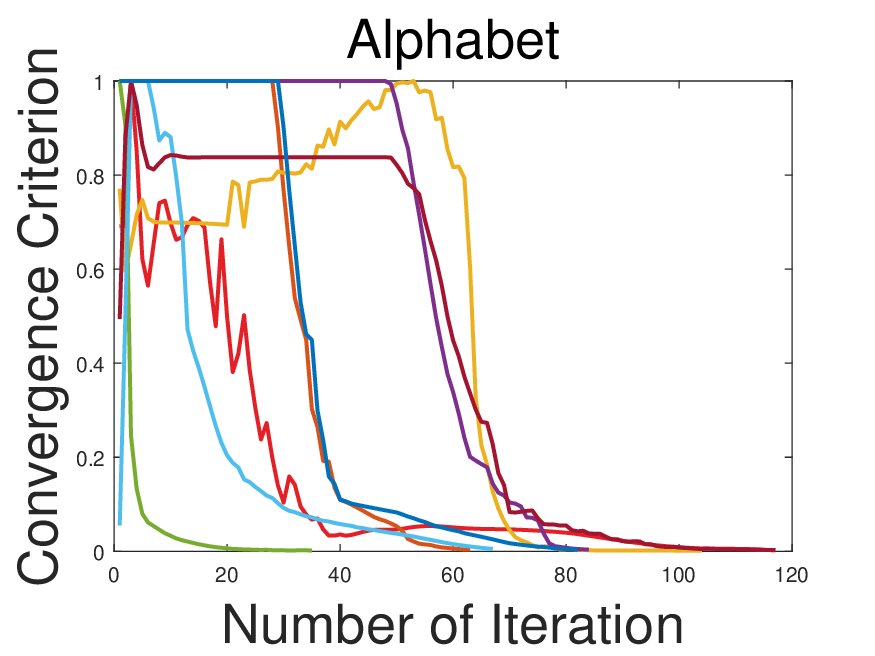}
                \label{fig_11_converge}}
            \subfloat[]{\includegraphics[width=3.8cm]{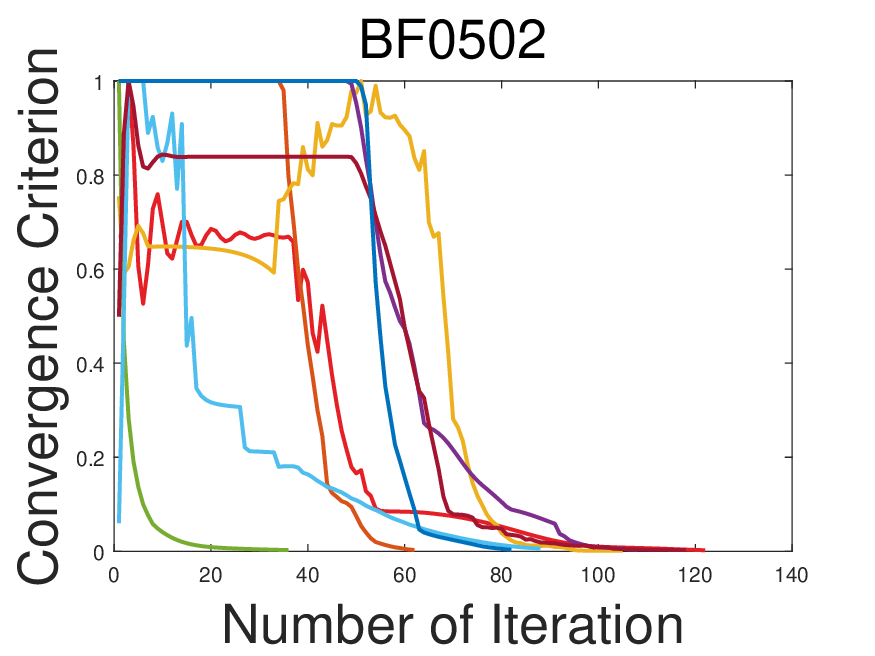}
                \label{fig_13_converge}}
          \subfloat[]{\includegraphics[width=3.8cm]{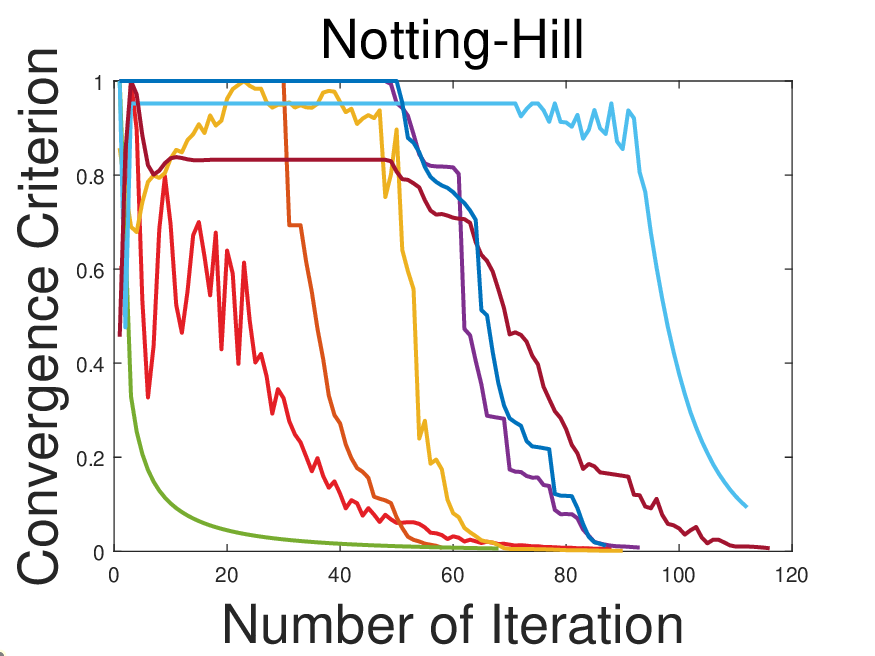}
        \label{fig_15_converge}}
            \caption{Convergence behavior comparisons of different methods on eight datasets. The longitudinal axis is normalized.}
            \label{fig_converge}
            \vspace{-0.04cm}
        \end{figure*}
	\begin{figure}[htbp]
            \vspace{-0.05cm}
            \centering
            \includegraphics[width=9cm]{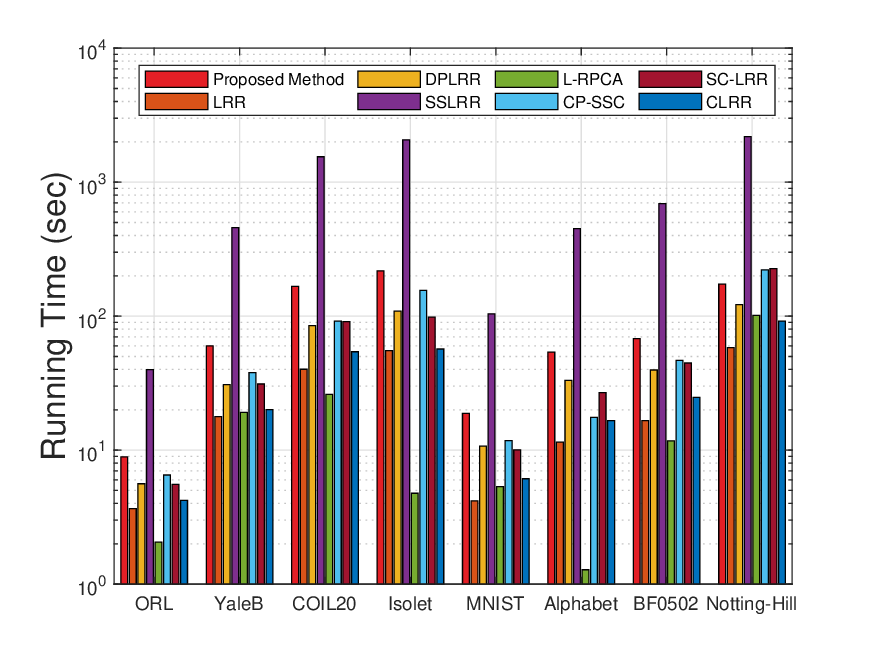}
            \caption{Running time comparisons of different methods on eight datasets.}
            \label{fig_time}
            \vspace{-0.4cm}
        \end{figure}
	
	In this section, we evaluated the proposed model on 8 commonly-used benchmark datasets, including ORL, YaleB, COIL20, Isolet, MNIST, Alphabet, BF0502 and Notting-Hill\footnote{ORL, YaleB, COIL20, Isolet and MNIST can be found in \url{http://www.cad.zju.edu.cn/home/dengcai/Data/data.html}, BF0502 in \url{https://www.robots.ox.ac.uk/~vgg/data/nface/index.html}, Notting-Hill in \url{https://sites.google.com/site/baoyuanwu2015/}.}.
	Those datasets cover face images, object images, digit images, spoken letters, and videos.
	To generate the weakly-supervisory information, following \cite{7931596}, we inferred the pairwise constraints from the randomly selected labels.
	
	We compared our method with LRR \cite{6180173} and six state-of-the-art semi-supervised subspace clustering methods, including DPLRR \cite{9454509}, SSLRR \cite{7931596}, L-RPCA \cite{9250019}, CP-SSC \cite{8682314}, SC-LRR \cite{6750063} and CLRR \cite{7726045}.
	We performed standard spectral clustering in \cite{8587179} on all the methods to generate the clustering result.
	We adopted the clustering accuracy and normalized mutual information (NMI) to measure their performance. 
	For both metrics, the larger, the better.
	For fair comparisons, we carefully tuned the hyper-parameters of the compared methods through exhaustive grid search to achieve the best result.
	To comprehensively evaluate the different methods, for each dataset, we randomly selected various percentages of initial labels $(\{5 \%, 10 \%, 15 \%, 20 \%, 25 \%, 30 \%\})$ to infer the pairwise constraints.
	We used the same label information for all the compared methods in every case.
	To reduce the influence of the random selection, we repeated the experiments 10 times with the randomly selected labels each time, and reported the average performance.

	\subsection{Comparison of Clustering Accuracy}
	Figs. \ref{fig_1}-\ref{fig_2} compare the clustering accuracy and NMI of different methods under various numbers of pairwise constraints, and Table \ref{table_1} shows the clustering performance of different methods with 30\% initial labels of each datasets as the supervisory information. From those figures and table, we can draw the following conclusions.
	\begin{enumerate}
		\item With more pairwise constraints, all the semi-supervised subspace clustering methods generally perform better, which indicates the effectiveness of including supervisory information in subspace clustering.
		\item The proposed method outperforms the other methods significantly. For example our method improves the accuracy value from 0.61 to 0.78 on MNIST and the NMI value from 0.72 to 0.89 on YaleB when compared with the best companions. According to Table \ref{table_1}, the proposed method improves the average clustering accuracy of the best compared methods from  0.83 to 0.92. Moreover, the proposed method almost always achieves the best clustering performance with varied supervisory information. 
		\item The compared methods may be sensitive to different datasets (e.g., SC-LRR achieves the second-best performance on YaleB and MNIST, but performs relatively bad on ORL and COIL20), and sensitive to diverse clustering metrics (e.g., CP-SSC performs well in NMI but poorly in clustering accuracy). On the contrary, the proposed method is robust to distinct datasets and metrics.
	\end{enumerate}
	
	Besides, we visualized the affinity matrices learned by different methods on MNIST in Fig. \ref{fig_visual}, where it can be seen that our method produces more dense and correct connections, leading to the most salient block diagonal affinity.
	This is owing to the used global tensor low-rank regularization, and further explains the good clustering results reported in Figs. \ref{fig_1}-\ref{fig_2} and Table \ref{table_1}.

 \vspace{-0.2cm}
	\subsection{Hyper-Parameter Analysis}\label{subsection_PA}
	Fig. \ref{fig_3} illustrates how the two hyper-parameters $\lambda$ and $\beta$ affect the performance of our method on COIL20, MNIST and Alphabet.
	It can be seen that the proposed model is relatively robust to hyper-parameters around the optimal.
	To be specific, we recommend to set 
	$\lambda\!=\!0.01$ and $\beta\!=\!10$.

\vspace{-0.2cm}
    \subsection{Convergence Speed}
    Fig. \ref{fig_converge} shows the convergence behaviors of all the compared algorithms on all the datasets. Note that the convergence criteria of all the methods are the same, i.e., the residual error of a variable in two successive steps is less than $1\mathrm{e}\!-\!3$.
     We can conclude that L-PRCA usually converges fastest among all the methods. While the proposed algorithm also converges fast compared with other methods like SSLRR and DPLRR. Moreover, the proposed algorithm gets converged within 130 iterations on all the eight datasets.

    Fig. \ref{fig_time} compares the average running time of all eight methods on each dataset. Note that we implemented all the methods with MATLAB on a Windows desktop with a 2.90 GHz Intel(R) i5-10400F CPU and 16.0 GB memory. We can observe that the proposed method takes an average time of 95.97s each run on all the eight datasets. It is slightly higher than LRR, but significantly lower than SSLRR. We also need to point out that the proposed method performs much better than the compared methods in clustering performance.

\vspace{-0.2cm}
	\subsection{Ablation Study}
	We investigated the effectiveness of the priors/processes involved in our model by comparing the clustering accuracy of Eqs. \eqref{eq_origin_woL}-\eqref{eqn_repair}.
	The compared methods include two well-known element-wise semi-supervised subspace clustering methods SSLRR and CLRR.
	As Table \ref{table_ablation} shows, the results of Eq. \eqref{eq_origin_woL} outperform SSLRR and CLRR significantly on all the datasets, which demonstrates the advantage of the global tensor low-rank prior over the element-wisely fusion strategy.
	Besides, Eqs. \eqref{eq_origin} and \eqref{eqn_repair} improve the performance of the proposed model successively, which indicates that both the graph regularization and the post-refinement processing contribute to our model.

    \vspace{-0.2cm}
	\section{Conclusion}\label{section_5}
	We have proposed a novel semi-supervised subspace clustering model. 
	We first stacked the affinity matrix and pairwise constraint matrix to form a tensor, and then imposed a tensor low-rank prior on it to learn the affinity matrix and augment the pairwise constraints simultaneously.
	In addition to the global tensor low-rank term, we added a Laplacian regularization term to model the underlying local geometric structure.
	Furthermore, the learned affinity matrix was refined by the augmented pairwise constraints.
	The proposed model was formulated as a convex problem, and solved by ADMM.
	The experimental results demonstrated that our model outperforms other semi-supervised subspace clustering methods significantly.

In the future, we will investigate how to incorporate our work with the existing semi-supervised learning neural networks \cite{9794286,8961099,li2021contrastive,li2022twin}. For example, we can use the proposed pairwise constraint enhancement as a loss function to train the neural networks in an end-to-end manner. Moreover, we will improve our method by solving the noisy pairwise constraints problem. 

	\ifCLASSOPTIONcaptionsoff
	\newpage
	\fi

\end{document}